\documentclass[sigconf]{acmart}
\setcopyright{none}
\renewcommand\footnotetextcopyrightpermission[1]{}
\acmConference[ISSTA 2023]{ACM SIGSOFT International Symposium on Software Testing and Analysis}{17-21 July, 2023}{Seattle, USA}

\author{Xuanqi Gao}
\affiliation{
  \institution{Xi'an Jiaotong University}
  \city{Xi'an}
  \country{China}
}
\email{gxq2000@stu.xjtu.edu.cn}

\author{Juan Zhai}
\affiliation{
  \institution{Rutgers University}
  \country{United States}
}
\email{juan.zhai@rutgers.edu}

\author{Shiqing Ma}
\affiliation{
  \institution{Rutgers University}
  \country{United States}
}
\email{shiqing.ma@rutgers.edu}

\author{Chao Shen}
\affiliation{
  \institution{Xi'an Jiaotong University}
  \city{Xi'an}
  \country{China}
}
\email{chaoshen@mail.xjtu.edu.cn}

\author{Yufei Chen}
\affiliation{
  \institution{Xi'an Jiaotong University}
  \city{Xi'an}
  \country{China}
}
\email{yfchen@sei.xjtu.edu.cn}

\author{Shiwei Wang}
\affiliation{
  \institution{Xi'an Jiaotong University}
  \city{Xi'an}
  \country{China}
}
\email{shiwei.wang@stu.xjtu.edu.cn}

\usepackage{tikz}
\usepackage{amsmath}
\usepackage{filecontents}
\usepackage[flushleft]{threeparttable}
\usepackage{amsmath,amsfonts}
\usepackage{graphicx}
\usepackage{textcomp}

\usepackage[normalem]{ulem}

 \usepackage{url}

\usepackage{lipsum,tabularx}

\usepackage{multicol}
\usepackage{multirow}

\usepackage{colortbl}
\usepackage{booktabs}
\usepackage{setspace}

\hypersetup{
  linkcolor={blue!70!black},
  citecolor={red!70!black},
  urlcolor={blue!70!black}
}
\def\Snospace~{\S{}}

\usepackage[T1]{fontenc}

\usepackage{algorithm}
\usepackage{algorithmicx}
\usepackage[noend]{algpseudocode}

\usepackage{balance}

\usepackage{bm}
\usepackage{fp}
\usepackage{siunitx}
\usepackage{amsthm}

\usepackage[labelfont=bf,font=small,skip=5pt]{caption}
\usepackage{subcaption}

\usepackage{comment}

\usepackage{enumitem}
\usepackage{fancyhdr}
\usepackage{framed}
\colorlet{shadecolor}{blue!20}

\usepackage{tikz}

\usepackage{xspace}

\newcommand{\boxbeg}{
  \vspace{2px}
  \noindent\begin{tabular}{|l|}\hline
    \begin{minipage}{3.2in}
      \vspace{2px}
      \noindent
      }

      \newcommand{\boxend}{
      \vspace{2px}
    \end{minipage} \\ \hline
  \end{tabular}
  \vspace{-10pt}
}

\AtBeginDocument{%
  \providecommand\BibTeX{{%
    \normalfont B\kern-0.5em{\scshape i\kern-0.25em b}\kern-0.8em\TeX}}}

\begin{document}

\newcommand{\sys}{\mbox{\textsc{Ciliate}}\xspace}

\title{\sys: Towards Fairer Class-based Incremental Learning by Dataset and Training Refinement}

\begin{abstract}
    Due to the model catastrophic forgetting problem, Deep Neural Networks (DNNs) need updates to adjust them to new data distributions.
    The common practice leverages incremental learning (IL), e.g., Class-based Incremental Learning (CIL) that updates output labels, to update the model with new data and a limited number of old data.
    This avoids heavyweight training (from scratch) using conventional methods and saves storage space by reducing the number of old data to store.
    But it also leads to poor performance in fairness.
    In this paper, we show that CIL suffers both dataset and algorithm bias problems, and existing solutions can only partially solve the problem.
    We propose a novel framework, \sys, that fixes both dataset and algorithm bias in CIL.
    It features a novel differential analysis guided dataset and training refinement process that identifies unique and important samples overlooked by existing CIL and enforces the model to learn from them.
    Through this process, \sys improves the fairness of CIL by 17.03\%, 22.46\%, and 31.79\% compared to state-of-the-art methods, iCaRL, BiC, and WA, respectively, based on our evaluation of three popular datasets and widely used ResNet models.
    Our code is available at~\cite{AnonymizedRepositoryAnonymousa}.
\end{abstract}



\maketitle

\section{Introduction}\label{sec:intro}

We envision that the Software 2.0 era which ships Artificial Intelligence backed software will enable more potential applications (e.g., auto-driving) and reshape our society~\cite{bojarskiEndEndLearning2016,huConvolutionalNeuralNetwork2014, kalchbrennerConvolutionalNeuralNetwork2014}.
Towards this direction, Deep Neural Network~(DNN) represented Machine Learning~(ML) techniques have shown their advantage over other solutions.
For example, virtual assistants are DNN-based applications that can understand natural language voice commands and complete tasks for the user~\cite{HeySiriOndevice}.
AI navigation is used by autonomous robotic systems in conjunction with deterministic lower-level control algorithms (e.g., PID controller)~\cite{palossi64mWDNNbasedVisual2019}.

ML models, including DNNs, train on a large volume of data and tries to generalize to unseen data.
In practice, the distribution of real-world data is typically hard to describe, if not impossible.
By following strict statistical principles and best practices, e.g., balancing the number of samples belonging to each class, we try to produce high-quality and representative training datasets.
Thus, training datasets are \textit{sampled}  data points from the real distribution, which cannot always accurately describe the real-world distribution.
When facing a new set of real-world data, the well-trained ML model typically needs updates to fit itself to the new data.
Moreover, data distribution shift is also common in practice.
For example, natural language processing~(NLP) deals with natural language artifacts that change over time as society evolves.
When a data distribution shift happens, the old model will have low accuracy.
That is, a model ages when exposed to new data or the data distribution shifts,
known as the \textit{model catastrophic forgetting} phenomenon.

Conventional model training typically requires a large training dataset and starts from randomized initial weights.
Updating the old model using this method asks for memorizing all training data from the beginning and all new data (if the update happens more than once, which is typically true).
This is time-consuming, storage inefficient, and environmentally unfriendly, especially because modern models are becoming larger and larger.
For example, GPT-2 requires 1 week of training on 32 TPUv3 chips~\cite{strubellEnergyPolicyConsiderations2019}.
Retraining these systems with each new data update has a substantial carbon footprint and is expected to increase over the next several years~\cite{sharirCostTrainingNlp2020}.

Researchers propose the Incremental Learning~(IL) pipeline to alleviate this problem.
For example, we perform Class-based Incremental Learning~(CIL) when the model needs to add more output labels because of the new data.
The basic idea of IL is to start from well-trained base models and update the model on new data.
To avoid forgetting the old knowledge, it also keeps a small-sized training dataset sampled from the old large training dataset.
During training, it leverages a parameter to balance the old and new knowledge, hoping that the new model can generalize to both old and new data distributions.

IL methods suffer from model bias problems, as documented by existing work~\cite{liLearningForgetting2018,kirkpatrickOvercomingCatastrophicForgetting2017,zhaoMaintainingDiscriminationFairness2020} and also our results (see \autoref{sec:eval}).
Through our analysis, we found that the bias problem is caused by both dataset bias and algorithm bias.
Specifically, the sampled dataset can be biased or imbalanced in the feature space.
Namely, the sampled dataset overemphasizes some features or is not able to cover all useful and unique features, which leads to the model learning biased features, and hence, the model makes biased predictions.
Existing gradient-based training algorithms, e.g., Stochastic Gradient Descent~(SGD), extravasate dataset bias.
They tend to find features that are easily differentiated (from the gradient aspect) from the rest of the features rather than identifying correct and robust features~\cite{katharopoulosNotAllSamples2018,agarwalEstimatingExampleDifficulty2022}.
Therefore, models trained by these algorithms are not realizable, especially when the dataset bias exists.
Existing methods focusing on IL fairness problems develop machine learning mechanisms that try to solve the bias problem.
For example, iCaRL~\cite{rebuffiIcarlIncrementalClassifier2017} improves the sampling method, BiC~\cite{wuLargeScaleIncremental2019} deals with the data imbalance problem, and WA~\cite{zhaoMaintainingDiscriminationFairness2020} adjusts weights to alleviate algorithm bias.
On one hand, these methods overlook the opportunities of leveraging software engineering techniques that can observe and analyze the root cause of the dataset and model bias problems to guide the IL fairness fix.
Moreover, they only provide partial solutions to one of the root causes that lead to IL bias issues.
As such, their fixes are superficial, and cannot fully address the issue.

Inspired by ethics-aware software engineering~\cite{brunSoftwareFairness2018,aydemirRoadmapEthicsawareSoftware2018,galhotraFairnessTestingTesting2017}, we present \sys, a novel incremental learning bias fixing framework (focusing on CIL) by refined datasets and training for deep neural networks.
The key idea of \sys is that it follows the best practice of software engineering by first observing and analyzing the root cause of a bias problem in a single IL incremental step.
Concretely, it performs a differential analysis on the base model and model trained with traditional IL methods to identify the overlooked samples in the dataset.
These samples carry unique but ``minor'' features in the training dataset.
When overlooked, the model makes biased predictions.
We refine the dataset by tagging each identified sample with a score that reflects the importance of this sample in solving the fairness problem.
The rest samples will not be affected.
To ensure that the model learns such samples as expected, we also refine the training procedure.
For normal samples (i.e., samples with low importance), we reuse existing IL methods as they can learn sufficient features from these samples already.
For samples tagged with high importance, we train them with random dropouts in each round so that different sets of neurons can learn the features.
Intuitively, this refined training approach enforces more neurons learning the unique features of the important samples.
When making predictions, the accumulated effect will correct biased predictions.

We built a prototype of \sys in Python and PyTorch.
Our results on CIFAR-100, Flowers-102, and Stanford Cars dataset show that \sys can effectively fix the bias IL models, and simultaneously increase model utility and fairness performance.

Our contribution can be summarized as follows.

\begin{itemize}[leftmargin=0.75cm]
    \item We propose a framework \sys to address the bias problem in CIL.
    It leverages a novel differential analysis to guide our new dataset and training refinement techniques that address dataset and algorithm bias problems in CIL.
    \item Our evaluation of CIFAR-100, Flowers-102, and Stanford Cars dataset show that \sys has superior performance to state-of-the-art methods, iCaRL, BiC, and WA.
    On average, the fairness performance CWV is improved by 17.03\%, 22.46\%, and 31.79\%, while the accuracy is improved by 11.75\%, 14.70\% and 1.56\% for iCaRL, BiC, and WA, respectively.
\end{itemize}

\section{Background and Motivation}\label{sec:bg}

\subsection{Incremental Learning}\label{sec:il}

Like source code artifacts, the intelligent component (e.g., DNNs) of Software 2.0 programs needs to be continuously updated to new data, to support continuous integration (CI) and continuous delivery (CD). 
For example, a DNN-based face recognition system used for employee authentication and recognition has to be updated when new employees come in.
However, updating the intelligent component remains a challenging task.
A straightforward approach is to redo the training process over all the old and new data.
It will cause high costs on time and computing resources, especially when nowadays the dataset and model scale up rapidly.
In addition, owing to security and privacy concerns, old data is not allowed to be stored for a long time in some cases, e.g., health care and financial services~\cite{lesortContinualLearningRobotics2020,mcclureDistributedWeightConsolidation2018}.
Another conventional approach is to incrementally fine-tune the model as long as the data shift is small, i.e., new data is similar to the training data.
But this fine-tuning method does not cope with large bulks of new data.
The model may suffer from catastrophic forgetting problems due to its inability to maintain the discriminative capability on previously seen data~\cite{mccloskeyCatastrophicInterferenceConnectionist1989,frenchCatastrophicForgettingConnectionist1999}.

\begin{figure}[tb]
    \centering
    \scalebox{0.9}{
    \includegraphics[trim={0.2cm 21.5cm 11.8cm 2cm},clip,width=\linewidth]{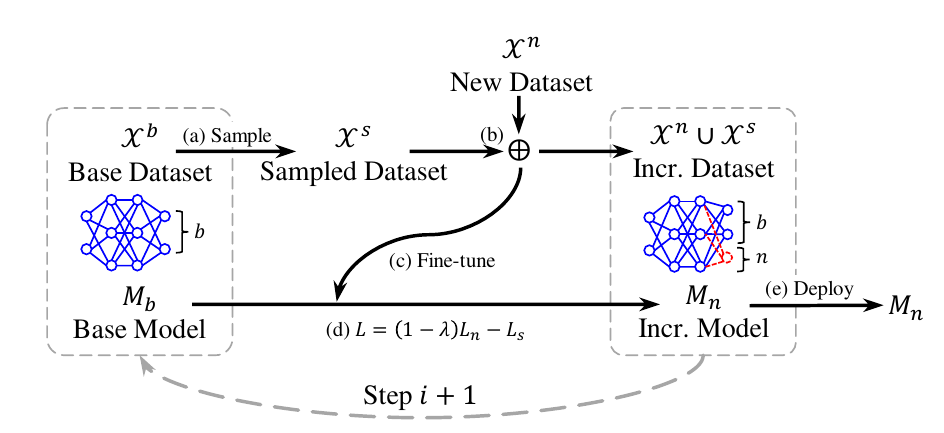}
    }
    \caption{Overview of the class-based incremental learning. \textit{Incr.} is short for \textit{Incremental}.}\label{fig:CIL}
    \vspace{-18pt}
\end{figure}

Incremental learning~(IL), also referred to as continual learning or lifelong learning, has been proposed to address the challenges of learning from a continuous stream of data.
It updates the model on a limited number of old data and new data and maintains a good accuracy for all data.
Therefore, it is more practical.
Depending on concrete scenarios, IL can have various types, including task-based IL, class-based IL, and domain-based IL~\cite{hsuReevaluatingContinualLearning2018}.
\autoref{fig:CIL} shows a simplified example, where we use a face recognition task to illustrate how class-based IL (CIL) works.
The base model \(M_b\) was trained on a large corps of facial data, \(\mathcal{X}^b\), belonging to \(b\) identities (i.e., the model has \(b\) output class labels).
With its expansion to new business, we want to extend the model to \(n\) more identities and the corresponding \(N\) facial data, \(\mathcal{X}^n\).
CIL memorizes a sampled dataset~(or exemplar) \(\mathcal{X}^s\) from the old training dataset \(\mathcal{X}^b\), i.e., \(\mathcal{X}^s \subset \mathcal{X}^b,\, |\mathcal{X}^s| \ll |\mathcal{X}^b| \) and fine-tunes the base model \(M_b\) on \(\mathcal{X}^n \cup \mathcal{X}^s\) with the following loss:
\begin{equation}\label{eq:combinedloss}
    L=(1-\lambda)L_n+\lambda L_s
\end{equation}
where \(L_n\) and \(L_s\) respectively are the cross-entropy losses (usually with temperatures) for the dataset \(\mathcal{X}^n\) and \(\mathcal{X}^s\), and \(\lambda\) is a parameter governing the balance between the two losses.
In short, this loss views the training on the old and new datasets as a multitask training process and uses a hyperparameter to balance these two tasks.
As discussed earlier, IL is a continuous process, formed by a series of incremental steps.
After finishing one incremental step, the new dataset \(\mathcal{X}^n \cup \mathcal{X}^s\) will be treated as the new base dataset, sampled based on hyperparameter \(\lambda\).
In the following incremental steps, the same process will repeat (on different datasets and labels) as illustrated in \autoref{fig:CIL}.

\begin{figure*}[t]
    \centering
    \footnotesize
    \scalebox{0.7}{
    \begin{subfigure}[t]{0.065\textwidth}
            \centering     
            \footnotesize
            \includegraphics[width=\textwidth]{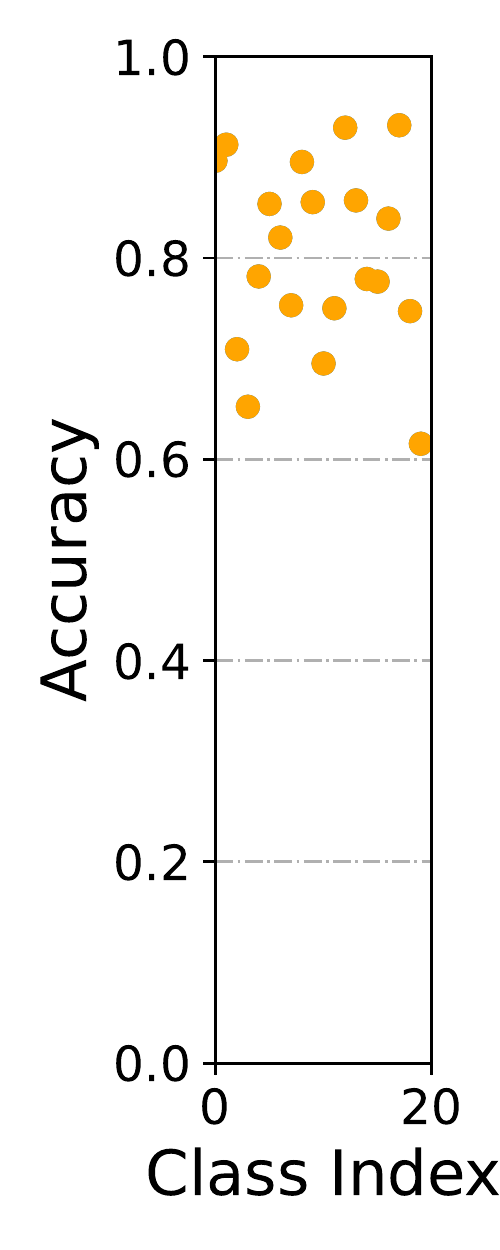}
                   \caption{\(M_b\)}\label{f:step1}
        \end{subfigure}
        \hfill
        \begin{subfigure}[t]{0.10\textwidth}
            \centering
                   \footnotesize
                   \includegraphics[width=\textwidth]{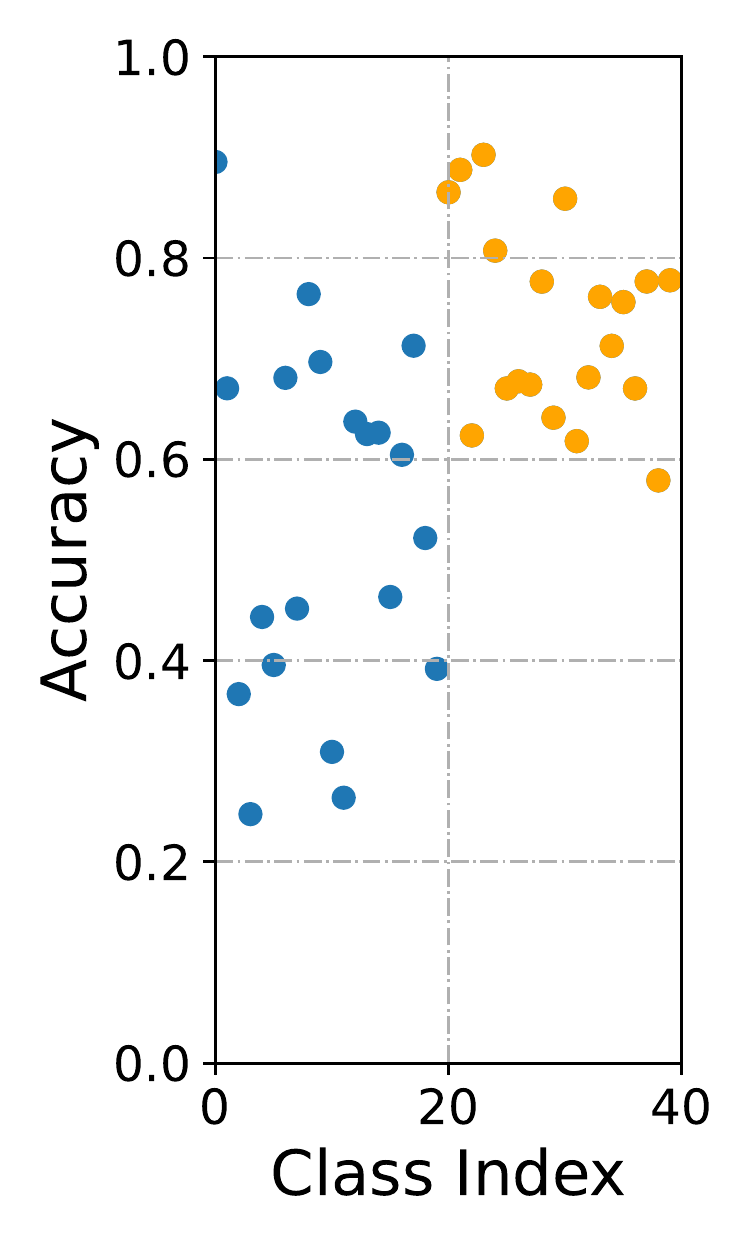}
                   \caption{Step 2}\label{f:step2}
        \end{subfigure}
        \hfill
        \begin{subfigure}[t]{0.13\textwidth}
            \centering     
            \footnotesize
            \includegraphics[width=\textwidth]{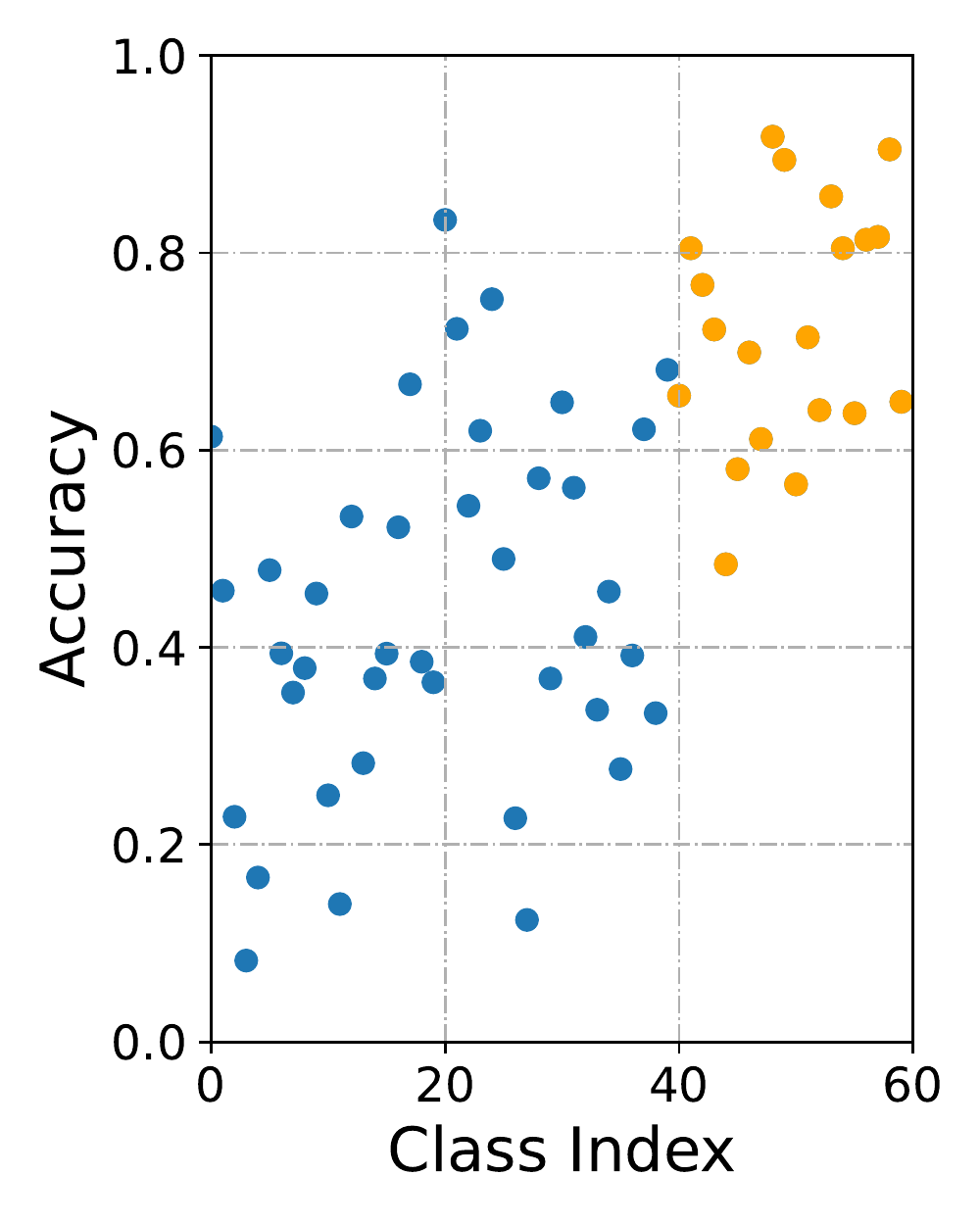}
            \caption{Step 3}\label{f:step3}
        \end{subfigure}  
        \hfill
        \begin{subfigure}[t]{0.165\textwidth}
            \centering     
            \footnotesize
            \includegraphics[width=\textwidth]{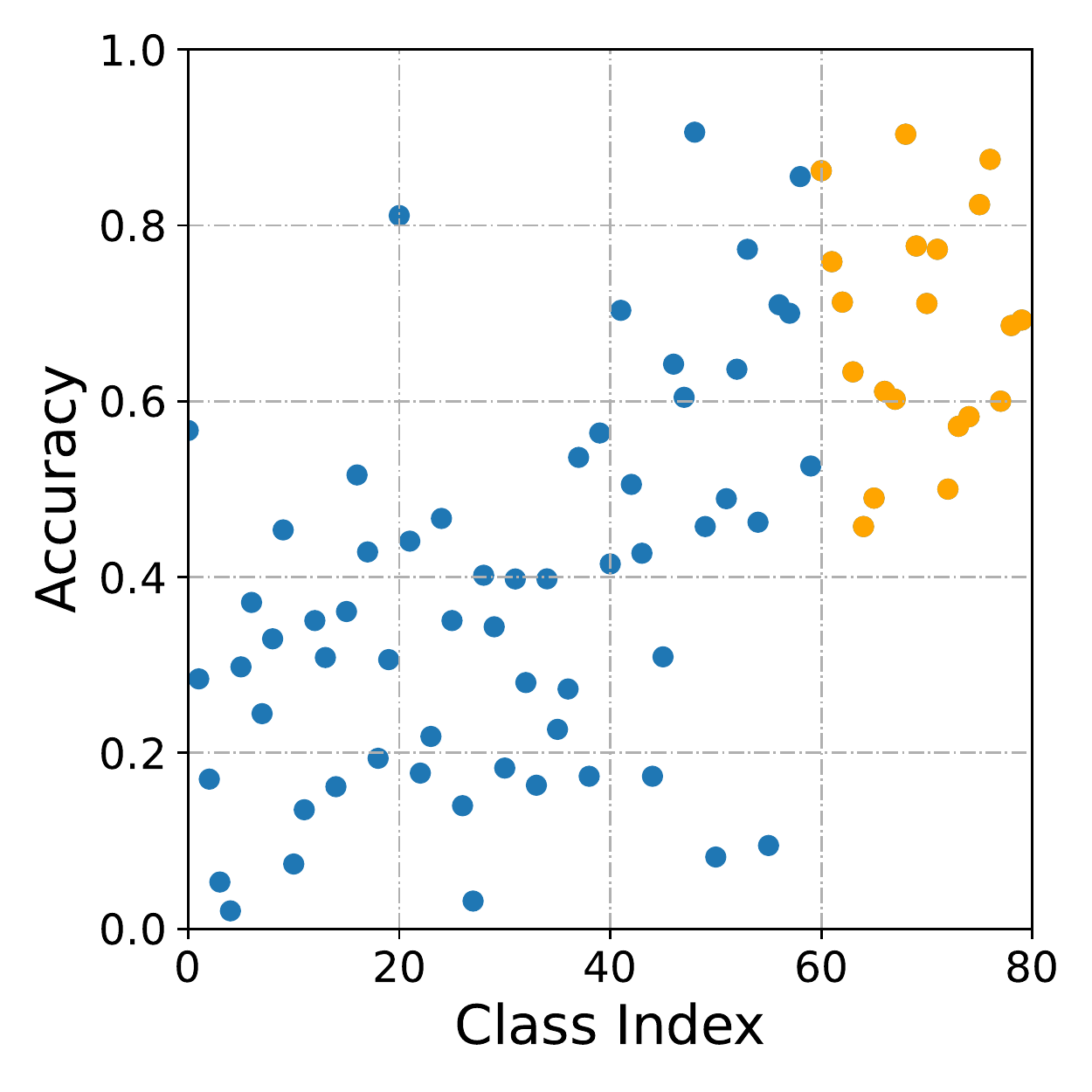}
                   \caption{Step 4}\label{f:step4}
        \end{subfigure}
        \hfill
        \begin{subfigure}[t]{0.20\textwidth}
            \centering     
            \footnotesize
            \includegraphics[width=\textwidth]{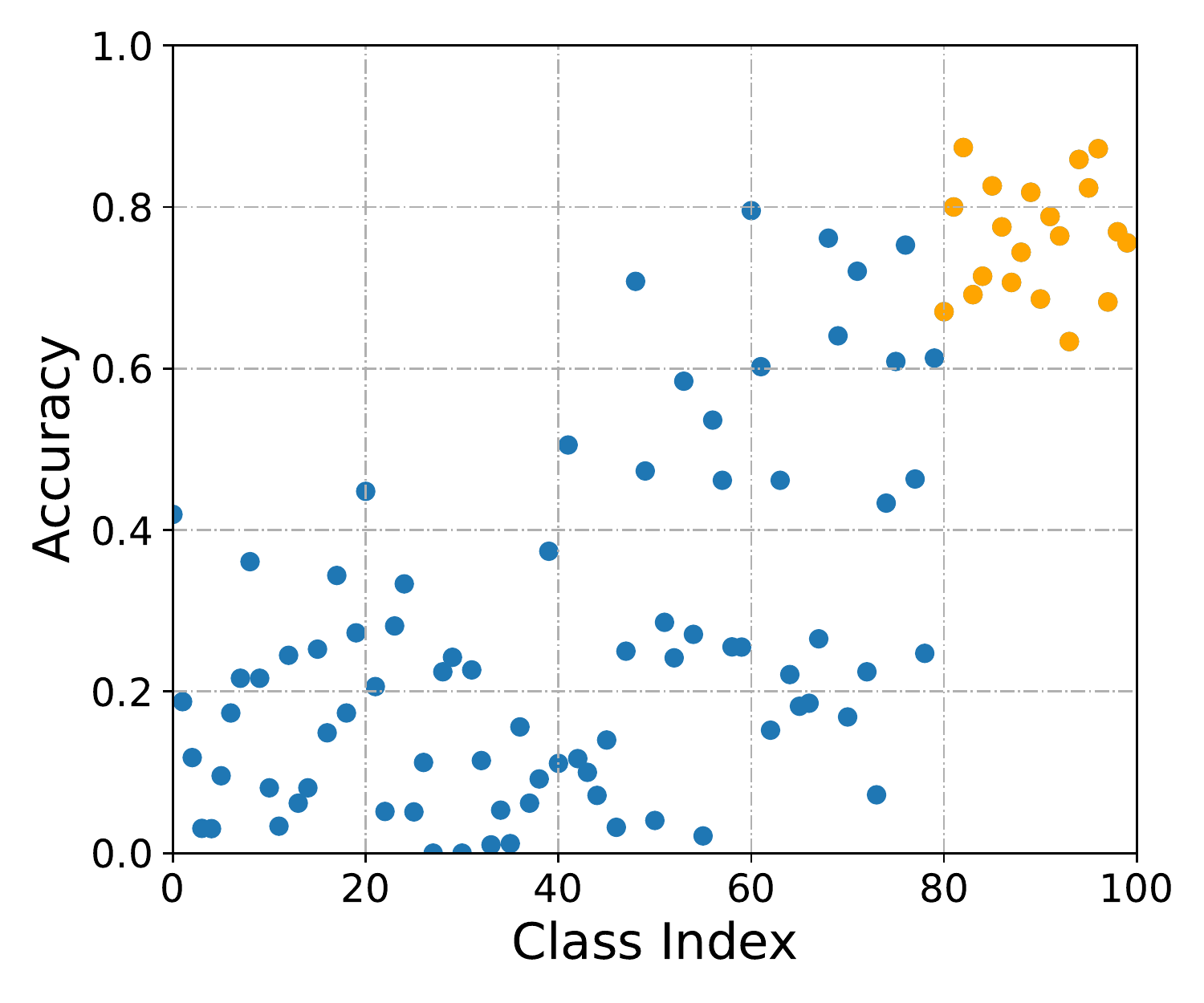}
            \caption{Step 5}\label{f:step5}
        \end{subfigure}
        \hfill
        \begin{subfigure}[t]{0.20\textwidth}
            \centering     
            \footnotesize
            \includegraphics[width=\textwidth]{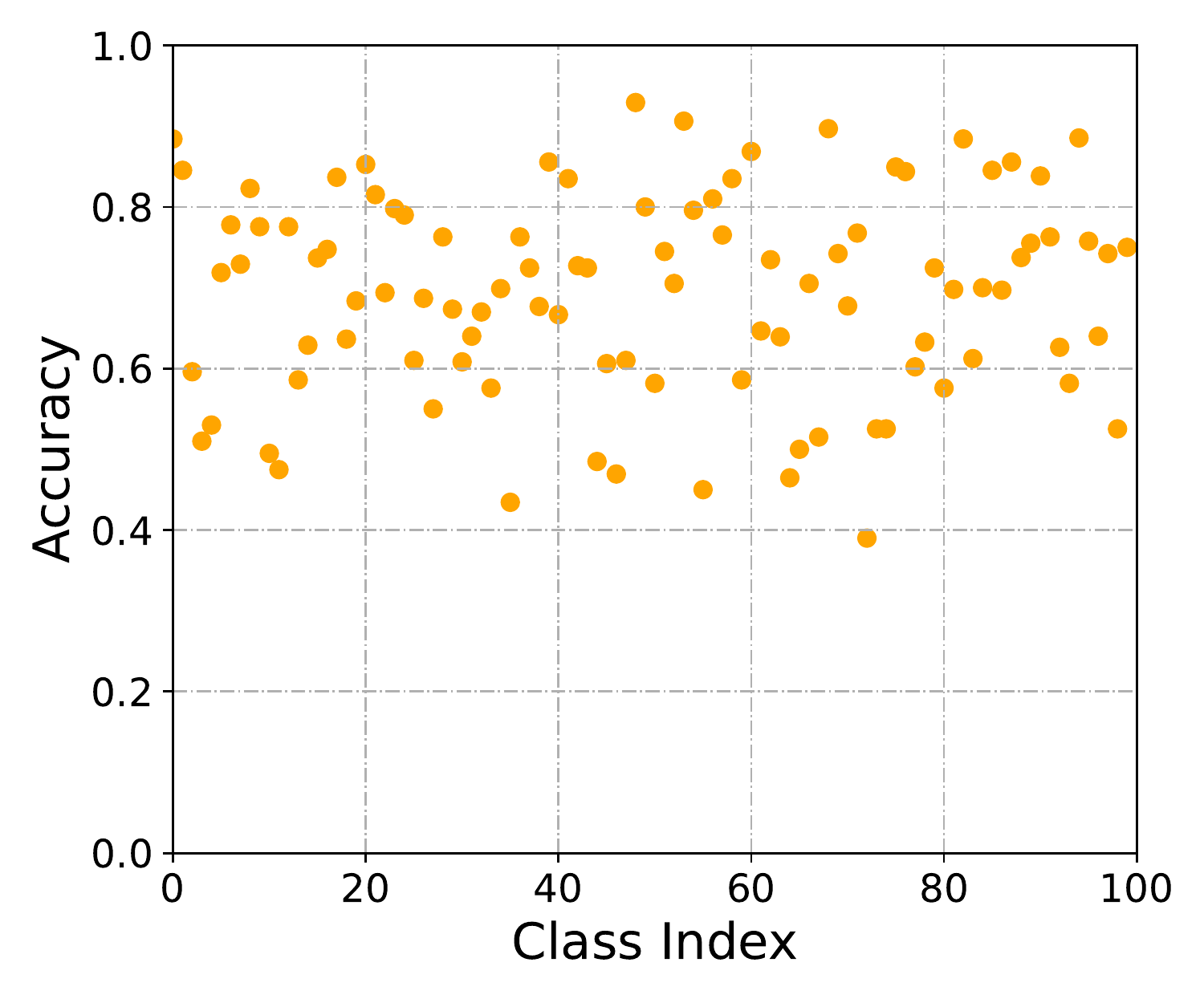}
                \caption{Traditional model}\label{f:stepall}
        \end{subfigure}
    }
    \caption{Class-wise accuracy of each step. (a) is the result of the base model, which does not correspond to incremental learning; (b)-(e) are the results of the $2^{nd}$ to $5^{th}$ incremental steps respectively; (f) is the result of the traditional non-incremental model obtained using all training samples for all classes. The orange points represent the new classes and the blue points present the old classes. The CWV of the model corresponding to each figure is 0.009, 0.030, 0.045, 0.056, 0.076, and 0.015, respectively.}\label{fig:class-scatter}
    \vspace{-10pt}
\end{figure*}

\subsection{Fairness of Incremental Learning}\label{sec:ilfairness}

In IL, the fairness of a model is typically measured by these metrics:

\smallskip
\noindent \textbf{Class-wise Variance~(CWV)~\cite{tianAnalysisApplicationsClasswise2021}.}
Given a dataset with \(C\) classes and the model accuracy of each class \(c\), \(a_c\), CWV can be defined as:
\begin{equation}\label{eq:cwv}
    \text{CV} = \frac{1}{|C|} \sum_{c\in C} ( a_c - \bar{a} )^2,\; 
    \bar{a}=\frac{1}{|C|}\sum_{c\in C} a_c
\end{equation}

\smallskip
\noindent \textbf{Maximum Class-wise Discrepancy~(MCD)~\cite{tianAnalysisApplicationsClasswise2021}.}
For a model, given the maximum and minimum class accuracy \(a_{\max}\) and \(a_{\min}\), respective, MCD can be defined as:
\begin{equation}\label{eq:MCD}
    \text{MCD} = a_{\max}-a_{\min}
\end{equation}

Intuitively, CWV measure the average discrepancy, and MCD uses the extreme discrepancy among classes, to estimate the fairness of a given model.
The larger these values are, the more biased the given model is.

IL typically suffers from fairness issues. 
To show this, we carry out experiments on CIFAR-100 with 5 incremental steps and 20 classes per step.
After each step, we calculate the accuracy of each class and plot them as scatter plots.
We also compare the fairness performance to the model obtained by conventional training, i.e., training a model with all classes and their training data.
\autoref{fig:class-scatter} shows how the accuracy varies (for each class) as the CIL progresses step by step.
Each point represents the accuracy of a single class, the orange ones represent the new classes, and the blue ones present the old classes.
As we can see, the accuracy of new classes is significantly better than those of old classes with a large variation, that is, the CIL model performs poorly in fairness.
In particular, the CWV of the model in each CIL training step is 0.009, 0.030, 0.045, 0.056, and 0.076, respectively.
MCD values are 0.32, 0.65, 0.83, 0.89 and 0.87, respectively.
Comparing the CIL model and the conventional training model that achieves 0.015 and 0.54 in CWV and MCD, we can tell that the CIL model is more biased.

\vspace{-10pt}
\subsection{Knowledge Distillation}\label{sec:kd}
Knowledge distillation~\cite{rebuffiIcarlIncrementalClassifier2017,zhaoMaintainingDiscriminationFairness2020,wuLargeScaleIncremental2019} is a technique used in machine learning to transfer knowledge from a complex, computationally expensive model (the teacher model) to a simpler, more lightweight model (the student model). 
The goal is to maintain the teacher model's performance while reducing the student model's computational requirements. 
In knowledge distillation, the student model learns to mimic the output of the teacher model, and the training objective is to minimize the difference between the output of the two models.
It is also commonly used to retain feature representations in incremental learning~\cite{zhaoMaintainingDiscriminationFairness2020,castroEndtoendIncrementalLearning2018,douillardPodnetPooledOutputs2020}.
\vspace{-10pt}

\section{System Design}\label{sec:design}

\subsection{Problem Statement}\label{sec:problem}

In this paper, we aim to develop a system that can automatically detect and fix fairness issues in class-based incremental learning.
Formally, for a given base model \(M_b\) and a new model \(M_n\) trained with class-based incremental learning, if \(f(M_n) - f(M_b) > \gamma \), where \(\gamma \) is a pre-defined threshold value based on concrete applications and \(f\) measures the fairness of the model, i.e., CWV or MCD, we say that the model has a \textit{fairness bug}.
Our system will be able to detect such bugs and fix them by training another model \(M_{\text{\sys}}\) by analyzing \(M_n\), \(M_b\), and their training data.
Following settings in prior methods~\cite{zhaoMaintainingDiscriminationFairness2020,wuLargeScaleIncremental2019}, we assume no control over the datasets \(\mathcal{X}^b\), \(\mathcal{X}^s\), and \(\mathcal{X}^n\).
Our method does not add new samples to datasets.
We control the training process, including how to use the samples in these datasets and train the model.

\subsection{Root Cause Analysis}\label{sec:analysis}
Model fairness is typically affected by two main factors: \textit{dataset bias}  and \textit{algorithm bias}.
By analysis, we found that both factors will affect the fairness of a model trained in CIL.

\subsubsection{Dataset bias}\label{sec:biasdataset}
Dataset bias is a common source of model bias~\cite{torralbaUnbiasedLookDataset2011,budaSystematicStudyClass2018}. 
Datasets can affect the model fairness from two aspects.
On the one hand, the imbalanced number of samples belonging to each class can lead to unfair decisions to disadvantage classes.
As an extreme example, a model learned from a million dog images and a single cat image can easily ignore the cat image (treat it as an outlier) and still achieve high precision by predicting all samples as dogs.
On the other hand, the quality of data can also affect the model's fairness.
Datasets with enough diverse features provide a complete set of information for the model to learn.
Otherwise, the model may only pick up a limited number of features and make predictions based on these unreliable features, leading to low fairness.
Our analysis shows that CIL cannot guarantee the quality of the new training data, leading to biased models.

CIL limits the number of old data, \(\mathcal{X}^s\).
It is one of the benefits of using CIL but also leads to an imbalanced number of samples for each class.
Typically, data samples belonging to new classes will be more than these of old classes, which is one of the root causes of dataset bias.
Moreover, the dataset \(\mathcal{X}^s\) is randomly sampled.
From a statistical aspect, a sampled distribution will be more biased if the number of samples is not large enough~\cite{katharopoulosNotAllSamples2018}.
That is, the dataset \(\mathcal{X}^s\) itself can be biased when the size is not large enough.
It eventually results in the narrow distributions of old classes on the feature space~\cite{wuLargeScaleIncremental2019,rebuffiIcarlIncrementalClassifier2017}.
Namely, not all features will be covered by the dataset, \(\mathcal{X}^s\).
As a result, the prediction accuracy for old classes will be affected.
If the number of old features is not high enough, the model will be significantly biased due to such dataset bias~\cite{quadriantoDiscoveringFairRepresentations2019}.

\begin{figure}[h]
    \centering
    \scalebox{0.75}{
    \includegraphics[trim={0 220 0 10},clip,width=\linewidth]{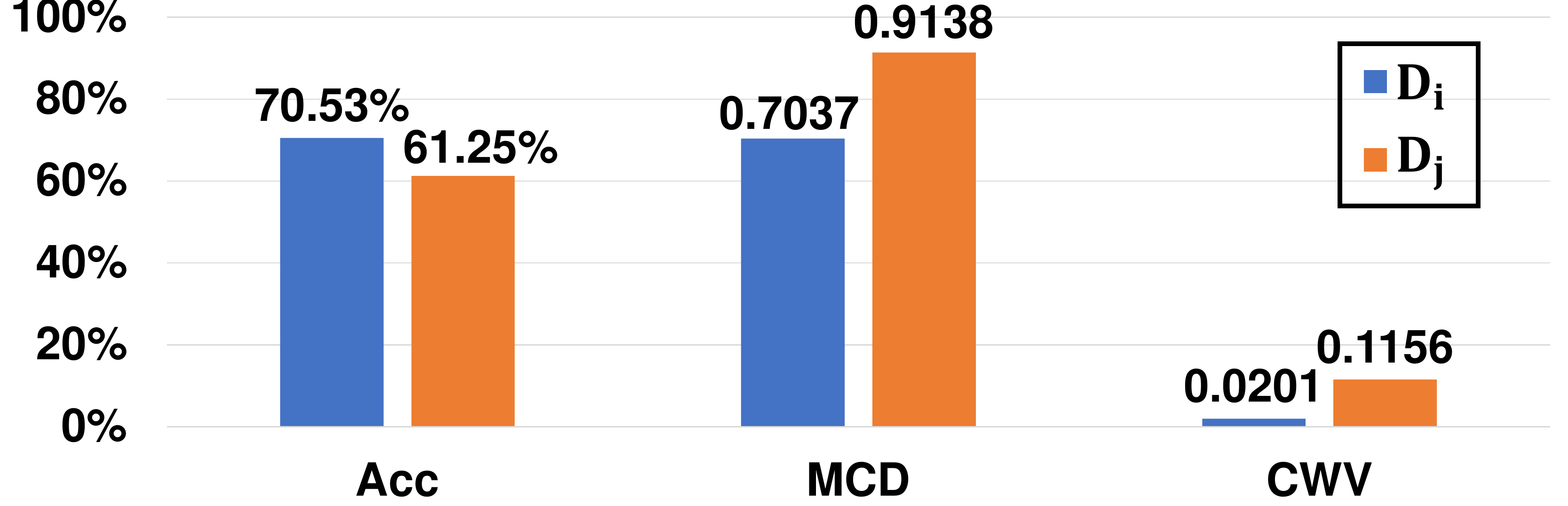}
    }
    \caption{Model performance on datasets with different settings: \(D_i\) with 10 labels, each containing 500 samples, and \(D_j\) with the same 10 labels, among which each class has 50 samples.}\label{fig:Imbalance}
    \vspace{-10pt}
\end{figure}

Firstly, we show the impact of dataset balance.
The experiment begins with a base model, using the ResNet-32 architecture~\cite{heDeepResidualLearning2016}, with 20 output labels, trained on 500 samples for each label.
We contrast CIL over two datasets, a dataset \(D_i\) with 10 labels, each containing 500 samples, and another dataset \(D_j\) with the same 10 labels, among which each class has 50 samples.
Then, we perform CIL using the same settings, i.e., 2000 samples as \(\mathcal{X}^s\) and the same set of hyperparameters, and obtain models \(M_i\) and \(M_j\) for dataset \(D_i\) and \(D_j\), respectively.
As we can see, dataset \(D_i\) is more balanced than the sampled dataset \(\mathcal{X}^s\) in terms of the number of training samples.
The accuracy, CWV, and MCD of the models are shown in \autoref{fig:Imbalance}.
The blue and orange bars denote \(M_i\) and \(M_j\), respectively.
As shown in the figure, \(M_i\) shows better fairness than \(M_j\), demonstrating the effect of dataset balance.

\begin{figure}[h]
    \centering
    \footnotesize
    \begin{subfigure}[t]{0.4\linewidth}
            \centering     
            \footnotesize
            \includegraphics[width=\textwidth]{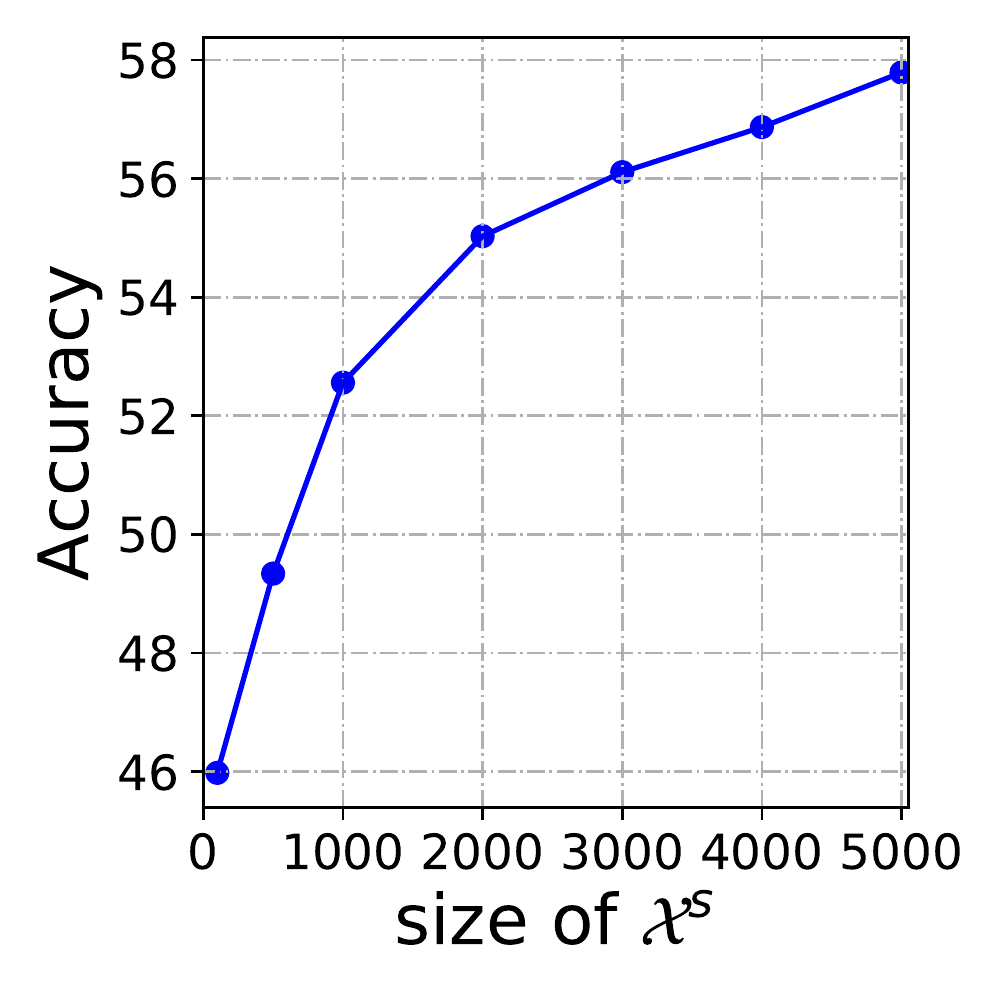}
                   \caption{Accuracy}\label{f:acc}
        \end{subfigure}
        \begin{subfigure}[t]{0.4\linewidth}
            \centering
                   \footnotesize
                   \includegraphics[width=\textwidth]{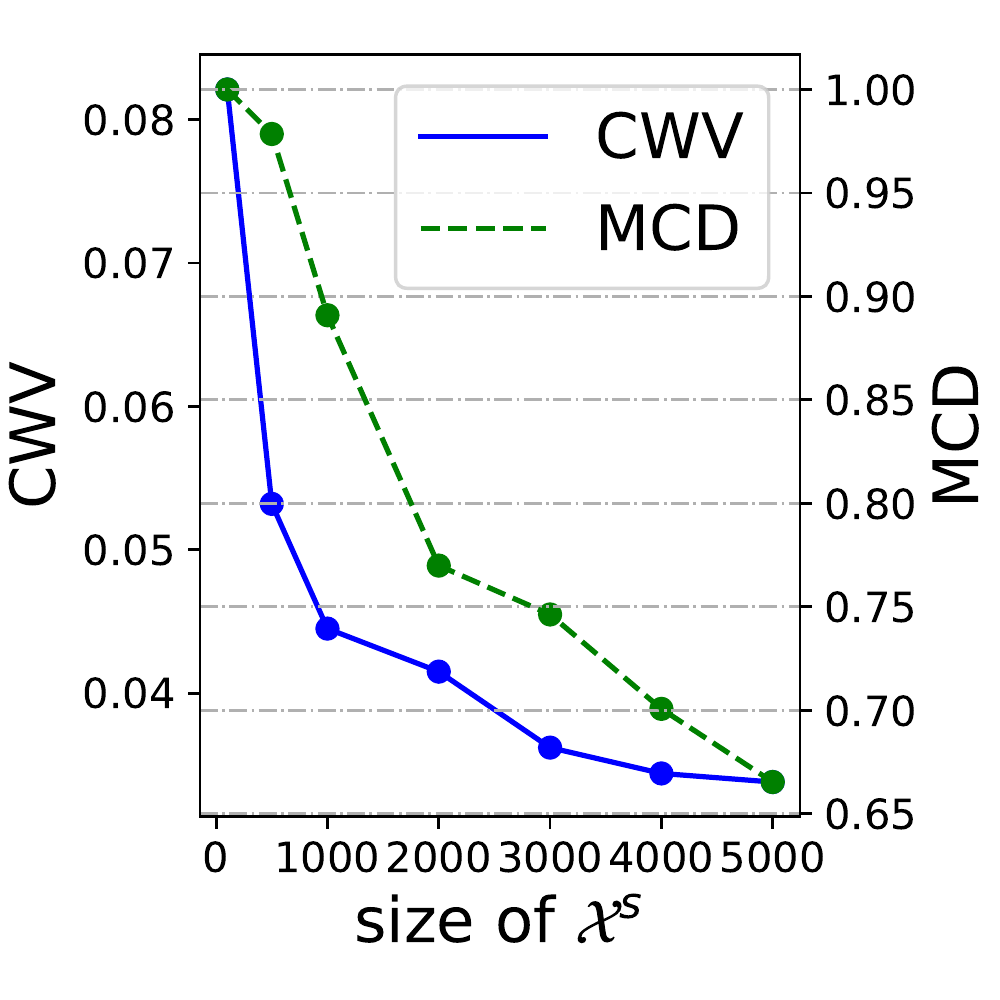}
                   \caption{CWV and MCD}\label{f:cwvmcd}
        \end{subfigure}

    \caption{Average model performance on CIFAR-100 with 20 classes per incremental step for different size of sampled dataset \(\mathcal{X}^s\).}\label{fig:memorybudget}
    \vspace{-10pt}
\end{figure}

Another experiment we do is to change the size of \(\mathcal{X}^s\) to see how its size affects the fairness of the model.
Intuitively, a larger size of \(\mathcal{X}^s\) means a larger number of old data, which will preserve more features in the old dataset.
Similar to the previous experiment, we use a ResNet-32 model trained on 20 output labels and 500 samples for each label as the base model \(M_b\).
Then, we use different sizes of \(\mathcal{X}^s\) and the same amount of new data, i.e., 20 output classes with each class containing 500 samples, to train new models using CIL.
We report the accuracy, CWV, and MCD of the models in \autoref{fig:memorybudget}.
As we can see, with the increase of \(\mathcal{X}^s\) size, the accuracy of the models is higher, while the CWV and MCD metrics are lower.
This means that the trained model benefits in both accuracy and fairness from training with a larger \(\mathcal{X}^s\) dataset.

\begin{figure}[h]
    \centering
    \scalebox{0.8}{
    \includegraphics[trim={0 220 0 0},clip,width=\linewidth]{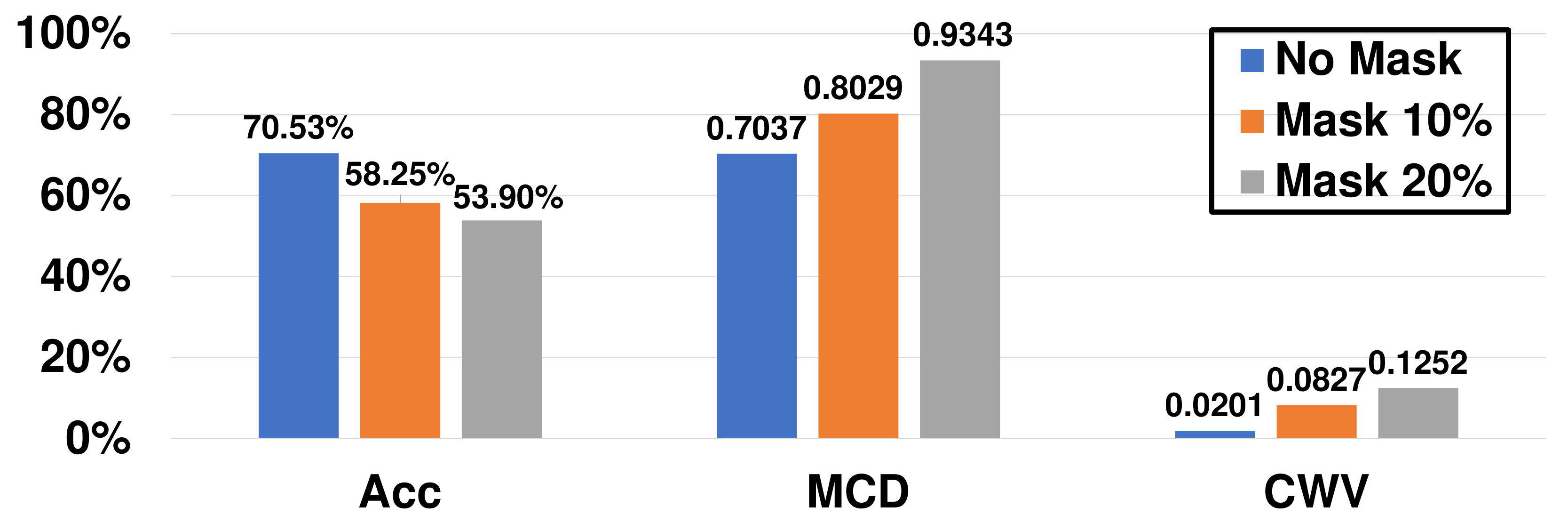}
    }
    \caption{Average model performance on CIFAR-100 dataset for different masking setup.}\label{fig:mask}
    \vspace{-10pt}
\end{figure}

Last, to verify that the number of features besides the number of samples affect the model fairness, we also conduct another experiment, start with a base ResNet-32 model trained on 500 samples that belong to 20 classes.
For the CIL step, the size of \(\mathcal{X}^s\) is 2000, the new data contains 20 output labels, and each label has 500 samples.
During training, we randomly mask input pixels to simulate the loss of input features.
The mask ratios \(\alpha\) are 0\%, 10\%, and 20\%.
\autoref{fig:mask} compares the accuracy, CWV, and MCD of these three models.
We use blue, orange, and gray colors to represent 0\%, 10\%, and 20\% of the mask ratios \(\alpha\), repsectively.
As we can see, with the loss of more features, the model becomes less accurate and less fair, showing the importance of the number of features.

To investigate whether randomly sampled data used by existing solutions can provide sufficient feature representation, we test the model neuron coverage during the CIL process.
Neuron coverage is widely adopted in DNN testing to guide the test generation for defect detection as a testing criterion~\cite{peiDeepXploreAutomatedWhitebox2017,xieDeepHunterCoverageguidedFuzz2019}.
Neurons can be regarded as a collection of input data features, so the neuron coverage reflects the feature representations of the model~\cite{erhanVisualizingHigherlayerFeatures2009}.
In this experiment, we start with a base ResNet-32 model trained on 500 samples that belong to 20 output labels.
For the CIL step, the size of \(\mathcal{X}^s\) is 2000, the new data contains 20 output labels, and each label has 500 samples.
We use the neuron coverage of the non-incremental learning model as the benchmark value.
If the difference between the coverage and this value is greater than 5\%, it is considered that an insufficient feature representation issue has occurred.
We conducted five rounds of experiments, and the experimental results are shown in \autoref{tab:nc}.
The first column lists the number of experiment runs.
The remaining columns show the neuron coverage of models in each step.
We have bolded values that do not meet the above criteria, and it can be seen that there are some issues of insufficient feature representation that occurred in the sampling process.

\begin{figure}[h]
    \centering
    \scalebox{0.65}{
    \includegraphics[trim={0 10 0 10},clip,width=\linewidth]{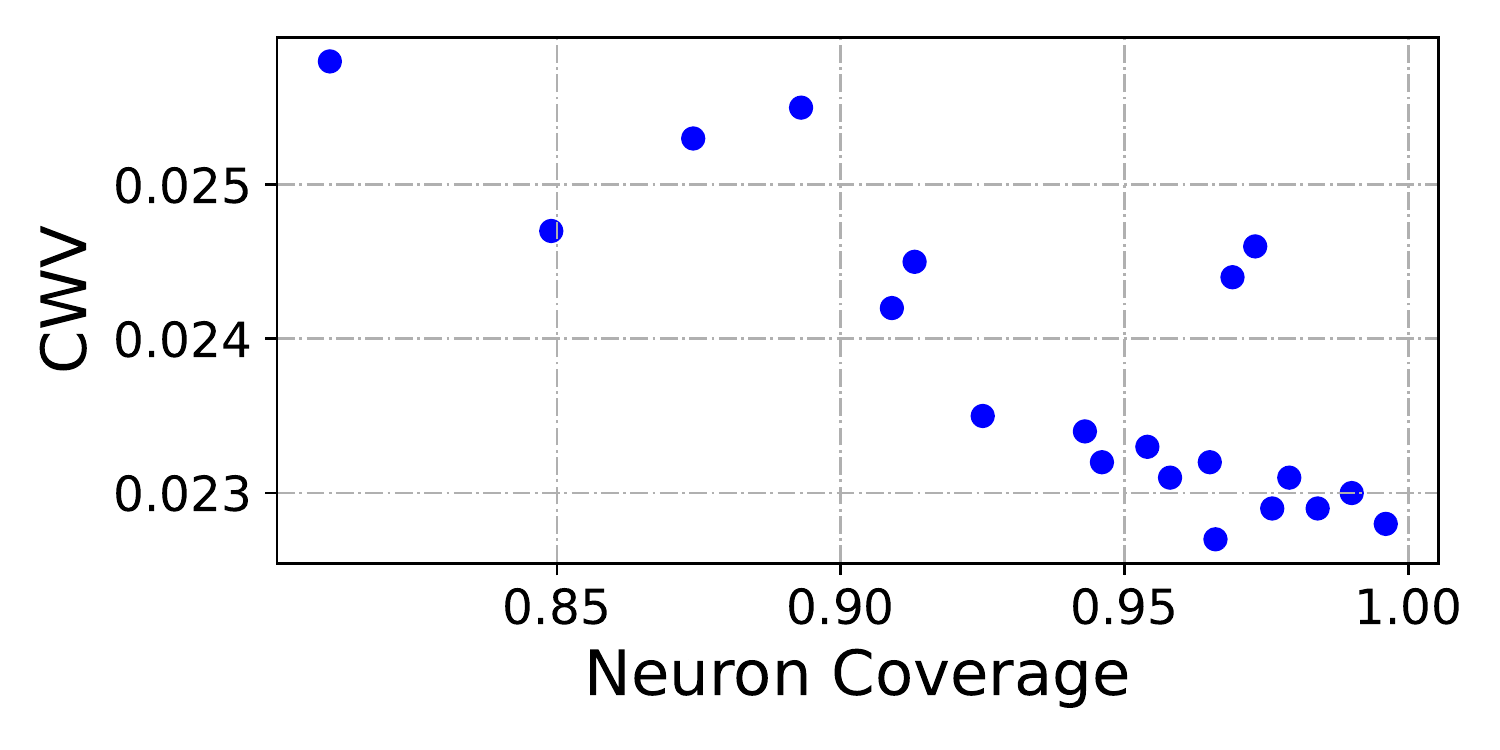}
    }
    \caption{Results of class-wise variance and neuron coverage in random sampling experiments. }\label{fig:ncbias}
    \vspace{-10pt}
\end{figure}
In this experiment, we start with a base ResNet-32 model trained on 500 samples that belong to 20 output labels.
For the CIL step, the size of \(\mathcal{X}^s\) is 2000, the new data contains 20 output labels, and each label has 500 samples.
We conducted 20 rounds of random sampling experiments, and the results are shown in \autoref{fig:ncbias}.
It can be seen that there is a negative correlation between neuron coverage and model bias, which indicates that the model with lower neuron coverage cannot obtain enough feature representation, which affects the performance of the model.

\begin{table}[]
    \centering
    \caption{Neuron coverage of model in CIL process. }
    \label{tab:nc}
    \scalebox{0.6}{
    \begin{tabular}{@{}lrrrrr@{}}
    \toprule
    Steps           & 1              & 2     & 3     & 4              & 5              \\
    \midrule
    Run 1           & 0.979          & 0.958 & 0.946 & \textbf{0.925} & \textbf{0.810} \\
    Run 2           & 0.975          & 0.959 & 0.966 & 0.986          & 0.991          \\
    Run 3           & \textbf{0.943} & 0.973 & 0.984 & 0.990          & 0.996          \\
    Run 4           & 0.966          & 0.977 & 0.980 & 0.994          & 0.998          \\
    Run 5           & 0.954          & 0.969 & 0.965 & 0.966          & 0.976          \\
    Non-incremental &                &       &       &                & 0.993    \\ 
    \bottomrule
    \end{tabular}
    }
    \vspace{-10pt}
\end{table}

From these experiments, we can conclude that dataset bias can significantly affect the fairness of the model trained by CIL.
A high-quality dataset shall reduce the dataset bias and emphasize the completeness of the data features.

\subsubsection{Algorithm bias}\label{sec:biastraining}
\begin{figure}[h]
    \centering
    \scalebox{0.7}{
    \includegraphics[trim={0 10 0 10},clip,width=\linewidth]{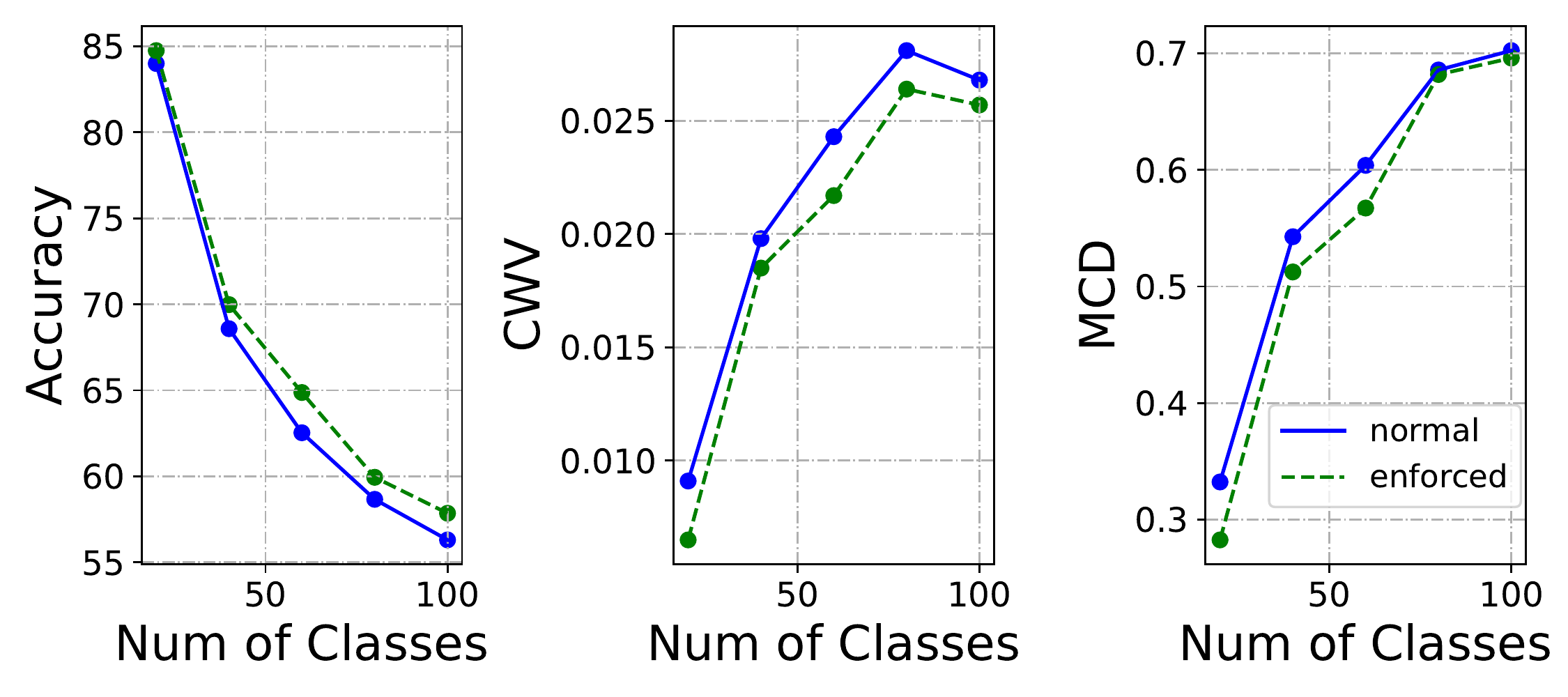}
    }
    \caption{Average model performance on a normal model and a model enforced to learn difficult samples by dropout training.}\label{fig:algbias}
    \vspace{-10pt}
\end{figure}

The learning process of the model may also introduce bias.
Firstly, the learning ability of the model on different samples is not the same.
Existing work~\cite{bengioCurriculumLearning2009,kumarSelfpacedLearningLatent2010} has shown that different samples have different learning difficulties for deep learning models, and existing gradient-based training algorithms tend to learn easier ones~(larger gradient values).
Samples with such essential yet hard-to-learn features are called difficult samples~\cite{kumarSelfpacedLearningLatent2010}.
Some models with high accuracy are highly biased because when optimizing during training, the model discards difficult features and hence, has lower fairness.
For CIL training, this effect is more pronounced, since optimizations are performed faster for new datasets which typically have smaller sizes.
Therefore, it is difficult for the model to maintain fairness during CIL training.
We compared the normally trained model and the model enforced to learn difficult samples.
In this experiment, we use the same base model as in the previous experiment, and we train the ResNet-32 model on a CIFAR-100 dataset with 20 output labels, each containing 500 samples.
Then, we prepare a set of samples with 20 output labels, each containing 500 samples, for a CIL step.
For one set of models, we train them with traditional CIL training.
For the comparison set, we select hard samples (will be discussed in \autoref{sec:dataset}) and enforce the model to learn their features by increasing the training iterations and leveraging dropout to train a larger set of neurons (see \autoref{sec:training}).
The accuracy, CWV, and MCD of these two sets of models are presented in \autoref{fig:algbias}.
The solid lines show the results of using traditional CIL training, and the dashed lines denote the new training.
The results show that we achieve higher accuracy and fairness in CIL tasks by enforcing the model to learn these difficult samples.
In particular, dropout-based training enforcement improve the model's accuracy by 2.22\%, CWV by 8.60\%, and MCD by 4.42\%.

\begin{figure}[h]
    \centering
    \scalebox{0.7}{
    \includegraphics[trim={0 10 0 10},clip,width=\linewidth]{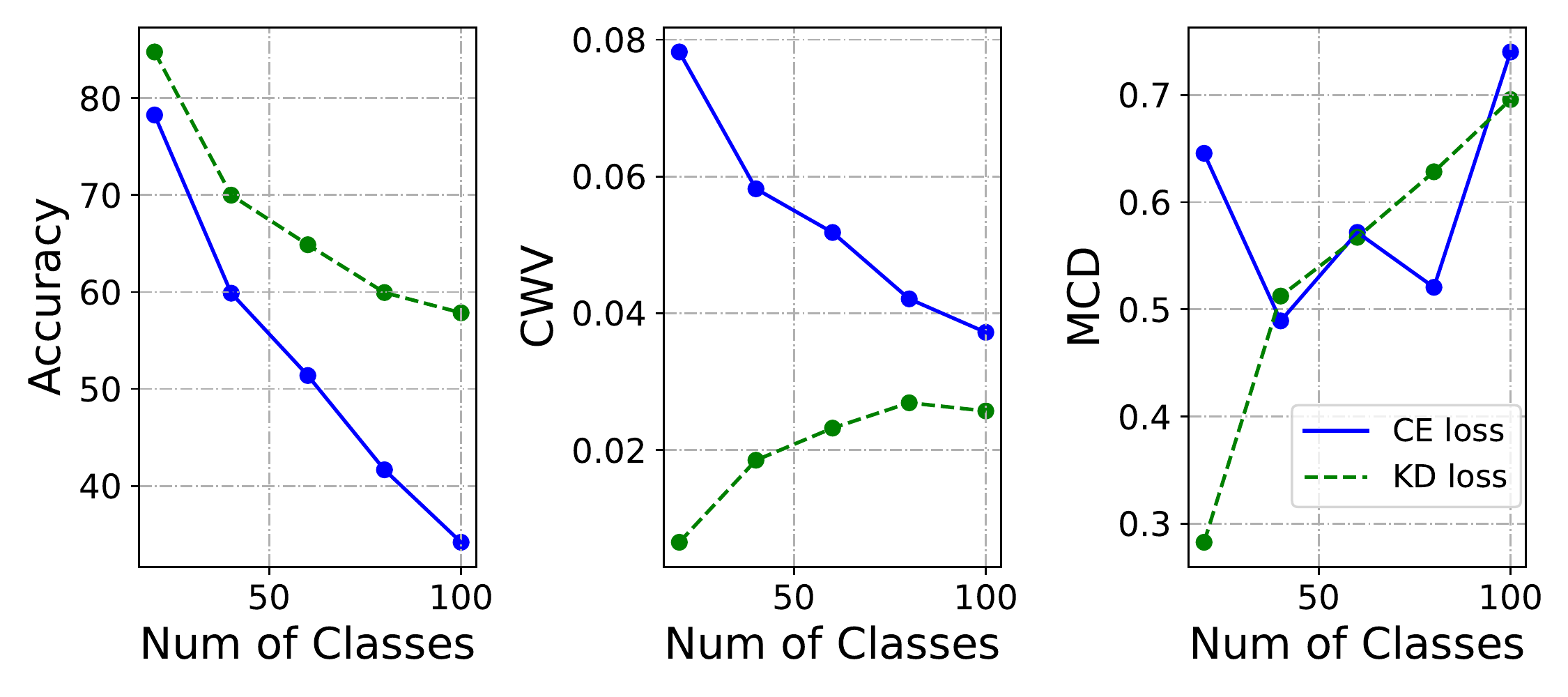}
    }
    \caption{Average performance on a model trained with normal cross-entropy loss and a model trained with knowledge distillation.}\label{fig:kdbias}
    \vspace{-20pt}
\end{figure}

Secondly, the data arrives in chronological order in CIL, which leads to the forgetting problem of the model, that is, the previous samples may be forgotten when the later samples are learned~\cite{douillardPodnetPooledOutputs2020,houLearningUnifiedClassifier2019}. 
Therefore, it is necessary to balance the plasticity~(ability to learn new classes) and rigidity~(ability to prevent forgetting) of the model.
To demonstrate this, we compare a model trained with normal cross-entropy loss and a model trained with knowledge distillation.
We maintain the same experimental setup as in the previous paragraph.
The results are presented in \autoref{fig:kdbias}.
The solid lines show the performance of the model trained with normal cross-entropy loss, and the dashed lines denote the model trained with knowledge distillation.
The results show that we achieve higher accuracy and fairness in CIL tasks by leveraging knowledge distillation.
In particular, the knowledge distillation model has been improved by 27.14\% of accuracy, 62.32\% of CWV, and 9.47\% of MCD.
\vspace{-10pt}

\subsection{System Overview}\label{sec:overview}

\begin{figure*}[h]
    \centering
    \scalebox{0.85}{
    \includegraphics[trim={1.5cm 16cm 1.5cm 7.8cm},clip,width=\textwidth]{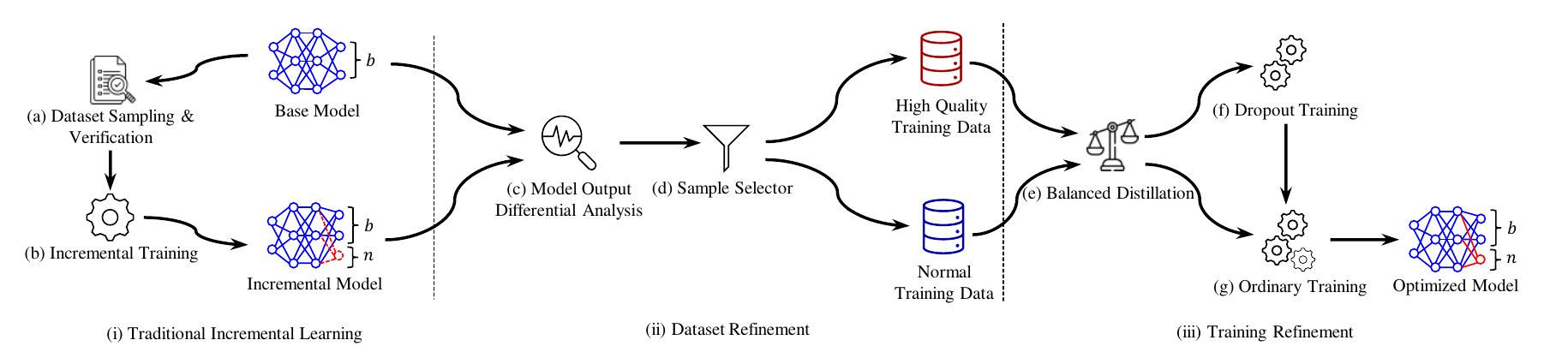}
    }
    \caption{One incremental step of {\sys}.}\label{fig:overview}
\end{figure*}

Based on our analysis, we know that to fix model fairness issues, we need to fix both dataset bias and also algorithm bias in CIL.
In \sys, we verify the data sampling process to avoid dataset bias caused by the under-representation of features, and we propose novel \textit{dataset refinement} and \textit{training refinement} to fix data and algorithm bias.
The goal of dataset refinement is to find a high-quality training dataset \(\mathcal{X}^h\), which identifies unique features that can be easily ignored by the CIL training.
And the training refinement will modify the training procedure to ensure that the minor data that carries unique and diverse features in \(\mathcal{X}^h\) will be well-learned so that the model eventually has a balanced performance.

Unlike traditional machine learning approaches~\cite{rebuffiIcarlIncrementalClassifier2017,zhaoMaintainingDiscriminationFairness2020,wuLargeScaleIncremental2019}, we leverage a software engineering approach.
\autoref{fig:overview} gives the overarching design for a single CIL incremental step in \sys.
First we apply traditional CIL on the base model \(M_b\) and get a new incremental model \(M_n\).
Then we conduct differential analysis on the outputs of the two models and use the results to select samples, thereby dividing the training data into two parts, \(\mathcal{X}^h\) and \(\mathcal{X}^l\).
Finally, we conduct different training methods on the two parts of data to get an optimized model \(M_c\) as the output of this incremental step.

The overall algorithm of \sys is presented in \autoref{alg:overview}, denoted as procedure \sys.
It takes a base model \(M_b\), a biased CIL model \(M_n\), and training data \(\mathcal{X}^t\) as inputs, and outputs a fixed model.
Firstly, \sys performs \textit{differential analysis} on model outputs of \(M_b\) and \(M_n\)~(line 1-6).
The basic idea is to analyze the model output changes before and after CIL and evaluate the importance of individual inputs to the two models, \(M_b\) and \(M_n\).
As traditional CIL trains on each sample with the same number of iterations, the few unique features will be easily ignored, which leads to bias.
By analyzing the differences in model outputs, we can estimate the unique features carried by each input sample, and identify the ones that vanished in the old CIL.
Based on these results, \sys performs dataset refinements, which highlights the important samples for fairness retraining, denoted as \(\mathcal{X}^h\)~(line 7).
The rest dataset is denoted as \(\mathcal{X}^l\).
Lastly, \sys performs refined training to enforce the buggy model to learn unbiased features~(line 8-12).
For those samples in \(\mathcal{X}^l\), we use the ordinary training methods, while for \(\mathcal{X}^h\), we conduct dropout training to enforce the buggy model to consider a larger set of features to avoid bugs.
By performing this refined training, \sys enforces the model to learn more unbiased features rather than biased ones to mitigate the fairness problem.

\begin{algorithm}[t]
\scriptsize
    \caption{\sys Algorithm}\label{alg:overview}
    \begin{algorithmic}[1]
      \Require \(M_b\): base model before incremental training
      \Require \(M_n\): biased incremental model obtained by incremental training
      \Require \(\mathcal{X}^t\): training dataset
      \Ensure \(M_c\): optimized model after fixing
      \Procedure{\sys}{}
      \State $L_d \gets []$
      \For{$s \in \mathcal{X}^t$}
      \State $P.div \gets JSLoss(M_b(s),M_n(s))$
      \State $P.sample \gets s$
      \State $Append(PathList,P)$
      \EndFor
      \State $\mathcal{X}^l,\mathcal{X}^h \gets GetRefined(L_d,\mathcal{X}^t,\eta)$
      \State $M_c \gets M_b$
      \State $E \gets GetErrorset(\mathcal{X}^t,M_n)$
      \For{$h \in \mathcal{X}^h$}
      \State $DropoutTraining(M_c,s)$
      \EndFor
      \For{$l \in \mathcal{X}^l$}
      \State $OrdinaryTraining(M_c,l)$
      \EndFor
      \Return $M_c$
      \EndProcedure

      \item[] 
      \Require \(L_d\): list of divergence corresponding to training samples
      \Require \(\mathcal{X}^t\): training dataset
      \Require \(\eta\): hyperparameter used to control the ratio
      \Ensure \(\mathcal{X}^l,\mathcal{X}^h\): Normal and high quality dataset, respectively
      \Procedure{GetRefined}{}
      \State $L_s \gets Sort(L_d.div)$
      \For{$i \gets 0 \  to \ \eta \times len(L_s)$}
      \State $Append(\mathcal{X}^h,L_s.sample)$
      \EndFor
      \For{$i \gets \eta \times len(L_s) \  to \ len(L_s)$}
      \State $Append(\mathcal{X}^l,L_s.sample)$
      \EndFor

      \Return $\mathcal{X}^l,\mathcal{X}^h$   
      \EndProcedure

    \end{algorithmic}

  \end{algorithm}

\subsection{Data Sampling}\label{sec:incremental}
At the beginning of the CIL pipeline, we follow traditional approaches to sample the base dataset to get a new dataset for incremental training~(\autoref{fig:CIL}(a)). 
As we show in \autoref{sec:analysis}, the sampled data may be under-represented, which can degrade training effectiveness and model fairness performance. 
Therefore, we added a re-sampling mechanism guided by neuron coverage.
More formally, given neurons of a model \(M\), \(|M|={n_1,n_2,\dots}\) and input data \(X={x_1,x_2,\dots}\), we have \(n(x)\), the output value of neuron \(n\) for a given input \(x\).
For a activation threshold \(t\), neuron coverage is defined as follows:
\begin{equation}\label{eq:NC}
    NC(M,\mathcal{X}) = \frac{|{n|\forall x\in \mathcal{X},n(x)>t}|}{|M|}
\end{equation}
In \sys, we feed the sampled dataset \(\mathcal{X}^s\) into the base model \(M_b\), and calculate the neuron coverage of \(M_b\).
If \(NC(M_b,\mathcal{X}^s) > \beta\), we consider the dataset well-sampled and directly use it for incremental training.
Otherwise, we resample the dataset.
We study the effect of parameters t and \(\beta\), and present the results in \autoref{sec:rq3}.
\vspace{-5pt}

\subsection{Dataset Refinement}\label{sec:dataset}

\subsubsection{Model Output Differential Analysis}\label{sec:differential}

The model output differential analysis is mainly to help \sys understand fairness bugs in the model, i.e., what samples are important for model fairness performance.
First, we feed input samples into models \(M_b\) and \(M_n\), and get their output values, denoted as \(M_b(s)\) and \(M_n(s)\), respectively.
Then, \sys calculates the Jensen-Shannon divergence value of model prediction on each sample~(\autoref{alg:overview} line 4).
Finally, we bind this divergence value to the sample and store them in a list~(denoted as \(L_d\)) for subsequent steps to use~(lines 5-6).

As we discussed in \autoref{sec:biasdataset}, the model bias is mainly introduced by the training dataset.
The divergence between the model outputs before and after training can reflect how much the input samples influenced the training of the base model \(M_b\).
That is, it reflects how important the input samples affect the model fairness.
Therefore, we iteratively calculate the divergence value of each input sample to obtain a criterion that can reflect the importance of the sample.
In mathematics, there are many ways to calculate divergence, including Kullback-Leibler divergence~\cite{kullbackInformationSufficiency1951}, Jensen-Shannon divergence~\cite{manningFoundationsStatisticalNatural1999}, and Hellinger divergence~\cite{hellingerNeueBegrundungTheorie1909}.
We test their performance and finally choose Jensen-Shannon divergence as the criteria~(see \autoref{sec:rq2} for a more detailed discussion).
The formula for calculating Jensen-Shannon divergence between distribution $P$ and distribution $Q$ is as follows:
\begin{equation}\label{eq:JS}
    JS(P,Q)=\frac{1}{2}\sum P\log(\frac{P}{M}) + \frac{1}{2}\sum Q\log(\frac{Q}{M})
\end{equation}
where \(M=\frac{1}{2}(P+Q)\).
We match the divergence value one-to-one with the samples after obtaining it in order to proceed to the next step, sample selection.

\subsubsection{Sample Selection}\label{sec:selection}
Sample selection's goal is to choose significant samples and integrate them into the dataset \(\mathcal{X}^h\).
The procedure \textit{GetRefined} in \autoref{alg:overview} shows the sample selection process, which separates the old training dataset \(\mathcal{X}^b\) into two datasets, \(\mathcal{X}^h\) and \(\mathcal{X}^l\).
First, we sort the \(L_d\) obtained above according to its divergence value to get an ordered list \(L_s\)~(line 15).
Then we add the samples corresponding to the parts with larger values in \(L_s\) to \(\mathcal{X}^h\), and the rest to \(\mathcal{X}^l\)~(line 16-19).
The cutoff point is determined by the product of the predetermined hyperparameter \(\eta\) and the length of \(L_s\), i.e. \(len(L_s)\).
Specifically, for the list \(L_s\) sorted in descending order, we calculate the cutoff point by the following formula:
\begin{equation}\label{eq:cutoff}
    P_c = \eta \times len(L_s)
\end{equation}
We designate any sample with an index below the cutoff point as an important sample, and we add it to \(\mathcal{X}^h\).
And for samples with an index greater than the cutoff point, we add them to \(\mathcal{X}^l\).

As was already indicated, the importance of each sample is reflected by \(L_d\).
By sorting \(L_d\), the importance of the samples is ranked naturally, allowing the most important samples to be distinguished.
We may change the size of the high-quality dataset \(\mathcal{X}^h\) by adjusting the hyperparameter \(\eta\), which changes the scale at which the model learns data features in the subsequent phase \textit{training refinement}.

\vspace{-5pt}
\subsection{Training Refinement}\label{sec:training}


\subsubsection{Balanced Distillation}\label{sec:distillation}
The training refinement aims to help \sys mitigate the algorithm bias introduced by model learning and forgetting~(discussed in \autoref{sec:biastraining}).
Inspired by PodNet~\cite{douillardPodnetPooledOutputs2020} and JTT~\cite{liuJustTrainTwice2021}, we propose an improved loss function, \textit{balanced distillation}.
As we discussed in \autoref{sec:biastraining}, a model adapted to new data tends to forget what it has learned previously. 
To this end, we introduce a rigid constraint through distillation loss \(L_r\), which helps the model maintain the previous knowledge~\cite{hintonDistillingKnowledgeNeural2015a}.
For those misclassified samples in the biased incremental model \(M_n\), we put them in the error set \(E\) and train them using cross-entropy loss \(L_{CE}\)~(\autoref{alg:overview} line 9). 
For correctly classified samples, \(L_r\) is used for training in order to retain this correct knowledge.
So the error set \(E\) can be formally expressed as follows:
\begin{equation}\label{eq:errorset}
    E = \left\{(x,y)|(x,y)\in \mathcal{X}^t \wedge M_n(x)\neq y\right\}
\end{equation}
Then the formula for the total loss function is as follows:
\begin{equation}\label{eq:totalloss}
    L = \lambda\sum_{(x,y)\in E}L_r(x) + \sum_{(x,y)\notin E}L_{CE}(x)
\end{equation}
We compare the results obtained by training with and without balanced distillation loss, as detailed in \autoref{sec:rq2}.
\vspace{-12pt}

\subsubsection{Dropout Training}\label{sec:dropout}

\begin{figure}[h]
    \centering
    \scalebox{0.9}{
    \includegraphics[trim={1.4cm 24cm 11cm 1.25cm},clip,width=\linewidth]{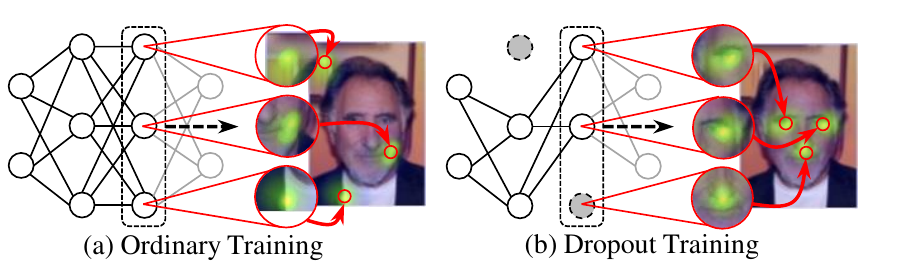}
    }
    \caption{Dropout technique example.}\label{fig:dropout}
    \vspace{-10pt}
\end{figure}

First, we set the base model \(M_b\) as the training starting point~(\autoref{alg:overview} line 8).
Then we conduct dropout training on \(\mathcal{X}^h\).
Finally, we conduct ordinary training \(\mathcal{X}^l\).
Note that both training procedures each contain multiple epochs of training steps.

Training refinement primarily applies the dropout technique when training on the \(\mathcal{X}^h\) dataset to help the model to learn a more fair representation and improve its fairness.
Dropout~\cite{srivastavaDropoutSimpleWay} is a regularization technique in which a certain percentage of nodes in a neural network are randomly ``dropped out'' (i.e., set to zero) during each training iteration. 
This forces the remaining nodes to learn more robust and independent features and helps to prevent overfitting, in which the model becomes too complex and memorizes the training data rather than learning from it. 
During testing, all nodes are used, but their outputs are scaled by the probability that they were retained during training so that the overall effect of dropout is still present. 
Dropout is a widely used technique in deep learning that has been shown to be effective in improving the generalization performance of neural networks on a wide range of tasks including alleviating fairness~\cite{websterMeasuringReducingGendered2020, gaoFairneuronImprovingDeep2022}.
Dropout training forces each neuron in the neural network to have the opportunity to learn a different representation of each sample, thus making difficult samples easier to be remembered and learned by the model, thereby mitigating the algorithm bias.

We also compare the results obtained by training without dropout or purely with dropout, as detailed in \autoref{sec:rq2}.

\section{Evaluation}\label{sec:eval}

We aim to answer the following research questions through our experiments:

\smallskip \noindent \textbf{RQ1:} 
How well can \sys fix biased models in CIL?

\smallskip
\noindent \textbf{RQ2:}
How do dataset refinement and training refinement affect the performance of \sys?

\smallskip
\noindent \textbf{RQ3:}
In \sys, what are the important tunable parameters/functions, and what are their effects?

\smallskip
\noindent \textbf{RQ4:}
What are the important lessons learned to help develop a better incremental learning pipeline?

\smallskip
\noindent \textbf{RQ5:}
How efficient is \sys in fixing biased models in CIL?

\subsection{Setup}\label{sec:setup}

\subsubsection{Hardware and software}
We conduct our experiments on a server with 64 cores Intel Xeon 2.90GHz CPU, 256 GB RAM, and 4 NVIDIA 3090 GPUs running the Ubuntu 16.04 operating system.

\subsubsection{Datasets}
We evaluate the methods on three popular datasets: CIFAR-100, Flowers-102~\cite{nilsbackAutomatedFlowerClassification2008}, and Cars~\cite{krause3dObjectRepresentations2013}.

\begin{itemize}[leftmargin=0.75cm]
\item \textbf{CIFAR-100:}
includes 60K 32$\times$32 RGB images of 100 classes. Each class has 500 training images and 100 testing images. 2000 samples are stored as exemplars.
\item \textbf{Flowers-102:}
includes 8189 RGB images of 102 classes. 
Each class consists of between 40 and 258 images. 
The training set and validation set each consist of 10 images per class (1020 images in total). 
The test set consists of the remaining 6149 images. 
1000 samples are stored as exemplars.
\item \textbf{Cars:}
contains 16,185 images of 196 classes of cars. 
The dataset is split into 8,144 training images and 8,041 testing images, where each class has been split roughly in a 50-50 split.
1000 samples are stored as exemplars.

\end{itemize}

\subsubsection{Models}
For a fair comparison with baseline methods, we reproduce our baselines, use the same models, and keep the same setup.
We train a 32-layer ResNet with SGD and set the batch size to 32.
The learning rate starts from 0.1 and reduces to 1/10 of the previous learning rate after 100, 150 and 200 epochs~(250 in total).
Random cropping, horizontal flip, and normalization are adopted for data augmentation.

\subsubsection{Baselines }
We compare \sys with other state-of-the-art class-based incremental learning methods, including iCaRL, BiC, and WA.

\smallskip
\noindent \textbf{iCaRL.}
As one of the early methods to introduce sampling in CIL, iCaRL proposes to sample old data based on data feature space representation~\cite{rebuffiIcarlIncrementalClassifier2017}.
Representations are extracted for all samples, and then each class's mean representation is calculated. 
The method iteratively selects exemplars for each class. 
At each step, an exemplar is selected so that, when added to the exemplars of a specific class, the updated exemplar centroid should be the closest to the real class centroid.

\smallskip
\noindent \textbf{BiC.}
Wu et al.~\cite{wuLargeScaleIncremental2019} proposed a bias correction method to mitigate the imbalance in incremental learning.
BiC adds an additional layer dedicated to correcting task bias to the network.
Then BiC divides a training session into two stages.
During the first stage, BiC trains the model consistent with the usual CIL.
Then in the second phase, BiC freezes all the parameters in the model and uses a split of a tiny part of the training data to serve as a validation set to learn the bias correction layer they added.

\smallskip
\noindent \textbf{WA.}
Zhao et al.~\cite{zhaoMaintainingDiscriminationFairness2020} proposed WA, which aligns the models' weights by adding new layers to improve the CIL performance.
They refine the bias weights in the FC layer following the regular CIL training, which helps the model adjust the biased weights.

These works improve the performance of CIL from the perspective of machine learning and enhance the generality of CIL.
However, they all raise fairness issues.
The reason is that none of them constrain the quality of the sampled dataset itself, nor use any metrics as guidance.
Thus, in fact, these methods conduct CIL training on a biased dataset, which makes it difficult for the model to achieve high fairness.

\subsubsection{Evaluation metrics}
We compare \sys with the baselines on three metrics: accuracy, CWV and MCD~(see \autoref{sec:il}).

\subsection{RQ1: Fixing Performance}\label{sec:rq1}

\noindent \textbf{Experiment Design:}
To evaluate the fixing performance of \sys, we test the following models: the naïve trained~(non-incremental) model, CIL models trained by BiC, by WA, by iCaRL and by \sys.
The naïve trained model is trained on the full dataset with the same trainer settings as the CIL model, as described in \autoref{sec:setup}.
For CIL model, the training begins with a base model~(step 1), with one-fifth of the dataset's overall label count since we conduct a 5-steps CIL.
Then for each incremental step, we train the model on a dataset consisting of the incremental dataset~(with one-fifth of the dataset's overall label count) and the sampled dataset~(updated based on the training dataset from the previous step while maintaining the same overall size).
After each incremental step, we evaluate the updated model on a subset of the test dataset, which contains samples related to the seen labels.
The size of sampled dataset for each dataset is mentioned in \autoref{sec:setup}.
We compare the performance between \sys and the other algorithms in terms of both utility and fairness.

\begin{table*}[]
    \centering
    \caption{Results on the three datasets. Best results are in bold.}
    \label{tab:effective}
    \scalebox{0.5}{
    \begin{tabular}{ll|rrrrr|rrrrr|rrrrr|rrrrr|rrrrr}
        \toprule
                                                   & \multicolumn{1}{c}{Incremental Step} & \multicolumn{5}{c}{1}                                                      & \multicolumn{5}{c}{2}                                                      & \multicolumn{5}{c}{3}                                                      & \multicolumn{5}{c}{4}                                                      & \multicolumn{5}{c}{5}                                                      \\
    \multicolumn{1}{c}{Dataset}                    & \multicolumn{1}{c}{Method}     & \multicolumn{1}{c}{Acc} & \multicolumn{1}{c}{Precision} & \multicolumn{1}{c}{Recall} & \multicolumn{1}{c}{CWV} & \multicolumn{1}{c}{MCD} & \multicolumn{1}{c}{Acc} & \multicolumn{1}{c}{Precision} & \multicolumn{1}{c}{Recall} & \multicolumn{1}{c}{CWV} & \multicolumn{1}{c}{MCD} & \multicolumn{1}{c}{Acc} & \multicolumn{1}{c}{Precision} & \multicolumn{1}{c}{Recall} & \multicolumn{1}{c}{CWV} & \multicolumn{1}{c}{MCD} & \multicolumn{1}{c}{Acc} & \multicolumn{1}{c}{Precision} & \multicolumn{1}{c}{Recall} & \multicolumn{1}{c}{CWV} & \multicolumn{1}{c}{MCD} & \multicolumn{1}{c}{Acc} & \multicolumn{1}{c}{Precision} & \multicolumn{1}{c}{Recall} & \multicolumn{1}{c}{CWV} & \multicolumn{1}{c}{MCD} \\ \midrule
    \multicolumn{1}{c}{\multirow{5}{*}{CIFAR-100}} & iCaRL                    & 81.95 & 79.83 & 79.77                & 0.0750                 & 0.4173                  & 69.85 & 68.13 & 68.09                & 0.0688                 & 0.5667                  & 63.07 & 61.56 & 61.49                & 0.0551                 & 0.6750                  & 57.30 & 55.87 & 54.82                & 0.0461                 & 0.7160                  & 55.03 & 52.89 & 52.34                & 0.0415                 & 0.7699                  \\
    \multicolumn{1}{c}{}                           & WA                       & \textbf{85.30} & 82.50 & 82.46       & 0.0726                 & 0.3788                  & \textbf{71.73} & 70.68 & 70.26       & 0.0837                 & 0.5385                  & \textbf{64.97} & 63.95 & 63.82       & 0.0784                 & 0.6814                  & 59.58 & 57.22 & 57.30                & 0.0699                 & \textbf{0.6403}         & 56.17 & 53.21 & 53.77                & 0.0641                 & 0.7365                  \\
    \multicolumn{1}{c}{}                           & BiC                      & 84.95 & 82.70 & 82.66                & \textbf{0.0691}        & 0.3833                  & 68.30 & 66.98 & 66.67                & 0.0740                 & 0.6322                  & 60.13 & 59.08 & 58.99                & 0.0657                 & 0.6618                  & 55.70 & 53.47 & 53.49                & 0.0619                 & 0.7589                  & 50.88 & 48.75 & 48.37                & 0.0532                 & 0.7343                  \\
    \multicolumn{1}{c}{}                           & \sys                      & 84.30 & 82.18 & 82.17                & 0.3214                 & \textbf{0.3178}         & 70.38 & 67.92 & 66.25                & \textbf{0.0177}        & \textbf{0.5141}         & 63.92 & 61.84 & 60.50                & \textbf{0.0216}        & \textbf{0.5821}         & \textbf{59.86} & 57.77 & 56.58       & \textbf{0.0262}        & 0.6453                  & \textbf{56.60} & 54.08 & 53.97       & \textbf{0.0256}        & \textbf{0.6888}         \\
    \multicolumn{1}{c}{}                           & Non-Incremental          &            --           &            --           &            --           &            --           &            --           &            --           &            --           &            --           &            --           &            --           &            --           &            --           &            --           &            --           &            --           &            --           &            --           &            --           &            --           &            --           & 70.53 & 67.21 & 66.33                & 0.0201                 & 0.7037                  \\ \midrule
    \multirow{5}{*}{Flowers}                       & iCaRL                    & 90.42 & 91.01 & 88.70                & 0.0969                 & 0.8000                  & 61.26 & 59.00 & 57.80                & 0.0985                 & 1.0000                  & 59.90 & 58.77 & 58.91                & \textbf{0.0877}        & 1.0000                  & 62.09 & 61.80 & 59.04                & 0.1307                 & 1.0000                  & 57.88 & 57.74 & 55.25                & 0.1146                 & 1.0000                  \\
                                                   & WA                      & \textbf{95.81} & 95.80 & 93.56       & 0.0865                 & 0.8000                  & \textbf{72.84} & 73.88 & 69.56      & 0.1289                 & 1.0000                  & 68.67 & 68.93 & 64.02                & 0.1417                 & 1.0000                  & 69.52 & 69.96 & 63.87                & 0.1505                 & 1.0000                  & 66.13 & 66.41 & 62.14                & 0.1423                 & 1.0000                  \\
                                                   & BiC                       & 93.75 & 93.69 & 91.77                & 0.0853                 & \textbf{0.7500}         & 62.89 & 63.92 & 60.01                & 0.1474                 & 1.0000                  & 61.02 & 61.37 & 59.98                & 0.1768                 & 1.0000                  & 60.72 & 61.13 & 59.73                & 0.1431                 & 1.0000                  & 55.46 & 55.82 & 53.66                & 0.1246                 & 1.0000                  \\
                                                   & \sys                      & 94.84 & 94.76 & 92.94                & \textbf{0.0662}        & \textbf{0.7500}         & 69.47 & 72.05 & 64.66                & \textbf{0.1058}        & 1.0000                  & \textbf{70.36} & 71.43 & 67.27       & 0.1044                 & 1.0000                  & \textbf{71.11} & 71.57 & 66.48       & \textbf{0.1211}        & 1.0000                  & \textbf{67.63} & 67.92 & 64.85      & \textbf{0.1103}        & 1.0000                  \\
                                                   & Non-Incremental          &            --           &            --           &            --           &            --           &            --           &            --           &            --           &            --           &            --           &            --           &            --           &            --           &            --           &            --           &            --           &            --           &            --           &            --           &            --           &            --           & 90.38 & 90.21 & 88.33                & 0.0933                 & 0.7500                  \\ \midrule
    \multirow{5}{*}{Cars}                          & iCaRL                    & 84.94 & 84.09 & 83.87                & 0.1581                 & 1.0000                  & 70.19 & 72.18 & 69.56                & 0.1579                 & 1.0000                  & 62.73 & 64.09 & 62.41                & 0.1374                 & 1.0000                  & 56.74 & 62.92 & 54.81                & 0.1349                 & 1.0000                  & 53.31 & 65.93 & 53.17                & 0.1274                 & 1.0000                  \\
    & WA                      & 87.98 & 87.24 & 87.00                & 0.1203                 & 0.7228                  & \textbf{74.67} & 74.33 & 73.80       & 0.1449                 & 1.0000                  & 64.71 & 72.39 & 63.74                & 0.1400                 & 1.0000                  & 60.15 & 70.78 & 59.83                & 0.1326                 & 1.0000                  & 60.59 & 72.60 & 60.05                & 0.1306                 & 1.0000                  \\
    & BiC                       & 88.08 & 87.48 & 87.10       & \textbf{0.1130}        & 0.8834                  & 73.54 & 72.61 & 71.77                & 0.1399                 & 1.0000                  & 64.25 & 69.03 & 63.62                & 0.1361                 & 1.0000                  & 59.40 & 68.85 & 58.79                & 0.1244                 & 1.0000                  & 55.61 & 69.02 & 55.07                & 0.1189                 & 1.0000                  \\
    & \sys                      & \textbf{88.24} & 87.61 & 87.06                & 0.1213                 & \textbf{0.6667}         & 72.30 & 71.92 & 70.66                & \textbf{0.1250}        & 1.0000                  & \textbf{65.60} & 76.78 & 64.14       & \textbf{0.1165}        & 1.0000                  & \textbf{61.23} & 75.74 & 59.94       & \textbf{0.1035}        & 1.0000                  & \textbf{63.12} & 76.94 & 60.44       & \textbf{0.1095}        & 1.0000                  \\
    & Non-Incremental          &            --           &            --           &            --           &            --           &            --           &            --           &            --           &            --           &            --           &            --           &            --           &            --           &            --           &            --           &            --           &            --           &            --           &            --           &            --           &            --           & 73.59 & 73.66 & 71.25                & 0.1133                 & 0.8077       \\ \midrule  
    \multirow{5}{*}{ImageNet}                          & iCaRL                    & 84.74 & 82.63 & 82.47                & 0.0232                 & 0.6014                  & 67.46 & 66.35 & 65.80                & 0.0217                 & 0.6487                  & 53.97 & 53.14 & 52.42                & 0.0197                 & 0.6833                  & 46.22 & 45.31 & 44.42                & 0.0209                & 0.7052                  & 42.75 & 42.11 & 41.25                & 0.0183                 & 0.7471                  \\
    & WA                      & 91.17 & 89.87 & 89.03                & 0.0227                 & 0.5735                  & 85.34 & 84.22 & 83.94       & 0.0214                 & 0.5990                  & 80.83 & 79.31 & 78.68                & 0.0222                 & 0.6419                  & 76.68 & 74.77 & 74.05                & 0.0213                 & 0.6637                  & 70.29 & 69.52 & 69.16                & 0.0204                 & 0.6915                  \\
    & BiC                       & 89.17 & 87.75 & 87.39       & 0.0215        & 0.5749                  & 84.35 & 81.97 & 81.82                & 0.0199                 & 0.6062                  & 78.09 & 77.14 & 76.86                & 0.0194                 & 0.6318                  & 73.67 & 71.79 & 71.28                & 0.0186                & 0.6721                  & 66.50 & 65.89 & 65.13                & 0.0178                 & 0.6825                  \\
    & \sys                      & 88.86 & 87.94 & 87.42                & 0.0322                 & 0.5311         & 84.45 & 82.17 & 81.73                & 0.0246        & 0.5571                  & 80.53 & 79.15 & 78.48       & 0.0215        & 0.5931                  & 76.47 & 75.16 & 75.25       & 0.0189        & 0.6152                 & 73.79 & 72.92 & 72.57       & 0.0181        & 0.6478                  \\
    & Non-Incremental          &            --           &            --           &            --           &            --           &            --           &            --           &            --           &            --           &            --           &            --           &            --           &            --           &            --           &            --           &            --           &            --           &            --           &            --           &            --           &            --           & 75.37 & 75.14 & 74.98                & 0.0109                 & 0.6953       \\ \bottomrule          
    \end{tabular}
    }
    \vspace{-10pt}
    \end{table*}
\noindent \textbf{Results:}
The comparison results are presented in \autoref{tab:effective}.
The first column lists the three datasets.
The second column shows the different algorithm.
The remaining columns list the model performance in each step, including accuracy~(Acc), class-wise variance~(CWV), and maximum class-wise discrepancy~(MCD).
The best results are highlighted in bold.
The experimental results demonstrate the effectiveness of our algorithm in fixing model fairness issues.
Firstly, \sys can effectively mitigate the fairness bias of class-based incremental learning.
Secondly, \sys achieves the highest utility among all algorithms.
\autoref{tab:effective} shows the fairness improvement of incremental learning on CIFAR-100, Flowers, and Cars, respectively.
\sys outperforms the state-of-the-art in terms of both the final and average incremental accuracy.
We also compare the model performance to the non-incremental model.

For fairness performance, \sys achieve the highest CWV among all models, which exceeds iCaRL by 33.26\%, 3.77\% and 14.07\%; WA by 56.75\%, 22.48\% and 16.13\%; and BiC by 47.94\%, 11.49\% and 7.94\% on CIFAR-100, Flowers and Cars, respectively.
\sys even surpasses the non-incremental model by 3.35\% on Cars dataset.
In terms of MCD, \sys exceeds iCaRL by 33.26\%, WA by 4.17\% and BiC by 3.89\% on CIFAR-100.
For Flowers and Cars dataset, there is no significant difference in their MCD performance.
We found that it is mainly because the distribution of these two datasets is more difficult to learn since the sampling size of old data is small, resulting in that there is always one or some classes that cannot be comprehensively learned.
Besides, \autoref{tab:effective} demonstrates that \sys has the advantage of increasing accuracy by successfully fixing fairness issues for CIL models.
The average utility of \sys outperforms all CIL models.
Compared to the non-incremental model, accuracy of \sys degrades 19.75\%, 25.18\% and 16.39\% on CIFAR-100, Flowers and Cars respectively, while iCaRL degrades 21.98\%, 35.96\% and 27.55\%; WA degrades 20.36\%, 26.84\% and 17.67\%; and BiC degrades 27.86\%, 38.64\% and 24.42\%.
On average, \sys improves CWV by 17.03\%, 22.46\% and 31.79\% compared to state-of-the-art methods, iCaRL, BiC, and WA.

\noindent \textbf{Analysis:}
From \autoref{tab:effective}, we make a few observations.
Firstly, compared with the other three CIL approaches, \sys is more effective~(higher test accuracy) and fair~(higher fairness performance).
In the meantime, the model accuracy is also improved after \sys fixing the fairness bugs, indicating that \sys does not degrade the effectiveness of the whole model.
Thus, we can say that \sys can effectively fix the bias CIL models, and simultaneously increase model utility and fairness performance.
\vspace{-10pt}

\subsection{RQ2: Impacts of Refinement}\label{sec:rq2}

\noindent \textbf{Experiment Design:}
In this section, we aim to demonstrate the effectiveness of the two core phases of \sys, the dataset and training refinement.
To this end,
1) for \textit{dataset refinement}, we compare the fixing performance among \sys, \sys without verification, and \sys without sample selection, that is, the random dataset refinement method where we set the size of the randomly-obtained refined dataset to be the same as that of \sys;
2) for \textit{training refinement}, we compare the fixing performance among \sys, \sys without balanced distillation, and \sys without selective training~(including pure dropout and pure ordinary training).
We follow the CIL setting in \autoref{sec:rq1} and conduct experiments on the CIFAR-100 dataset.
We compare these methods in terms of the average performance~(except the base model) and the performance of the model obtained in the last step.

\noindent \textbf{Results:}
\begin{table}[]
    \centering
    \caption{Comparison among the models trained by different components. The average performance~(except the base model) and the performance of the model obtained in the last step are reported.}
    \label{tab:ablation}
    \scalebox{0.55}{
    \begin{tabular}{@{}lrrrrrr@{}}
    \toprule
    Method & Avg Acc.  & Avg CWV & Avg MCD & Last Acc.  & Last CWV & Last MCD \\ 
    \midrule
    \sys  & 62.69    & 0.0228           & 0.6076          & 56.60    & 0.0256           & 0.6888  \\ 
    w/o Sample Selection  & 60.65  & 0.0261             & 0.6413 & 54.92    & 0.0287           & 0.7146            \\ 
    w/o Verification & 62.42  & 0.0233             & 0.6161 & 55.92    & 0.0278           & 0.7122            \\
    w/o Balanced Distillation & 60.88  & 0.0268       & 0.6662 & 53.93    & 0.0281     & 0.7139            \\
    Pure Dropout  & 61.60  & 0.0249             & 0.6214  & 55.73    & 0.0267           & 0.7009           \\ 
    Pure Ordinary  & 61.52  & 0.0248             & 0.6336 & 56.30    & 0.0268           & 0.7022            \\
    WA & 63.36 & 0.0740 & 0.6492 & 56.17 & 0.0641 & 0.7365 \\
    WA w/ Dropout & 63.29 & 0.0751 & 0.6411 & 56.22 & 0.0638 & 0.7298 \\
    BiC & 58.75 & 0.0637 & 0.6968 & 50.88 & 0.0532 & 0.7343 \\
    BiC w/ Dropout & 58.90 & 0.0644 & 0.7016 & 50.93 & 0.0540 & 0.7234 \\
    \bottomrule
    \end{tabular}
    }
\end{table}
The details of the comparison results are presented in~\autoref{tab:ablation}.
The first column lists the methods.
The next three columns show the average results over all incremental steps except the first step, and the remaining columns show the model performance in the last step.
Detailed results of all incremental steps are reported in the supplementary material.
Overall, \sys achieves the best fixing performance among the compared dataset and training techniques. 
It indicates that our proposed dataset and training refinement techniques are helpful to improve the fixing performance.

\textit{Impact of dataset refinement:}
The first three rows of \autoref{tab:ablation} present the results of the comparison.
With sample selection, the average accuracy has been improved by 2.90\%, and the fairness performance has also been improved, which are 9.66\% and 3.97\% of CWV and MCD, respectively.
With verification, \sys improves model performance by 0.43\%, 2.15\% and 1.38\% of accuracy, CWV and MCD, respectively.

\textit{Impact of training refinement:}
The first, fourth, fifth and sixth rows of \autoref{tab:ablation} present the results of different training refinement approaches.
Our balanced distillation improves model performance by 2.97\%, 14.93\% and 8.80\% of accuracy, CWV and MCD, respectively.
Our selective training method surpasses the ordinary training by 1.45\%, 4.03\%, and 3.97\%, while surpassing the pure dropout training by 1.31\%, 4.62\%, and 0.74\% in accuracy, CWV, and MCD respectively.
It confirms that selective training in \sys can achieve high accuracy and fairness.
The last four rows of \autoref{tab:ablation} present the impact of applying the dropout technique to BiC and WA.
Dropout changes WA performance by -0.11\%, -1.49\% and 1.25\%, while changing BiC performance by 0.26\%, -1.10\% and -0.69\% in accuracy, CWV, and MCD respectively.

\noindent \textbf{Analysis:}
From the results, we can draw some inspirations.
Firstly, compared with the other method, \sys achieves the highest utility and fairness performance, which proves the effectiveness of each phase in \sys.
Secondly, sample selection has a greater impact on performance, and verification has the least impact, which provides guidance for us to tune the parameters in \autoref{sec:rq3}.
Last, there is no significant performance improvement applying dropout to WA and BiC, indicating that dropout needs to be combined with dataset refinement to be effective.
\vspace{-10pt}
\subsection{RQ3: Effects of Tunable Configurations}\label{sec:rq3}
\begin{figure}[t]
    \centering
    \footnotesize
    \begin{subfigure}[t]{0.32\linewidth}
            \centering     
            \footnotesize
            \includegraphics[width=\textwidth]{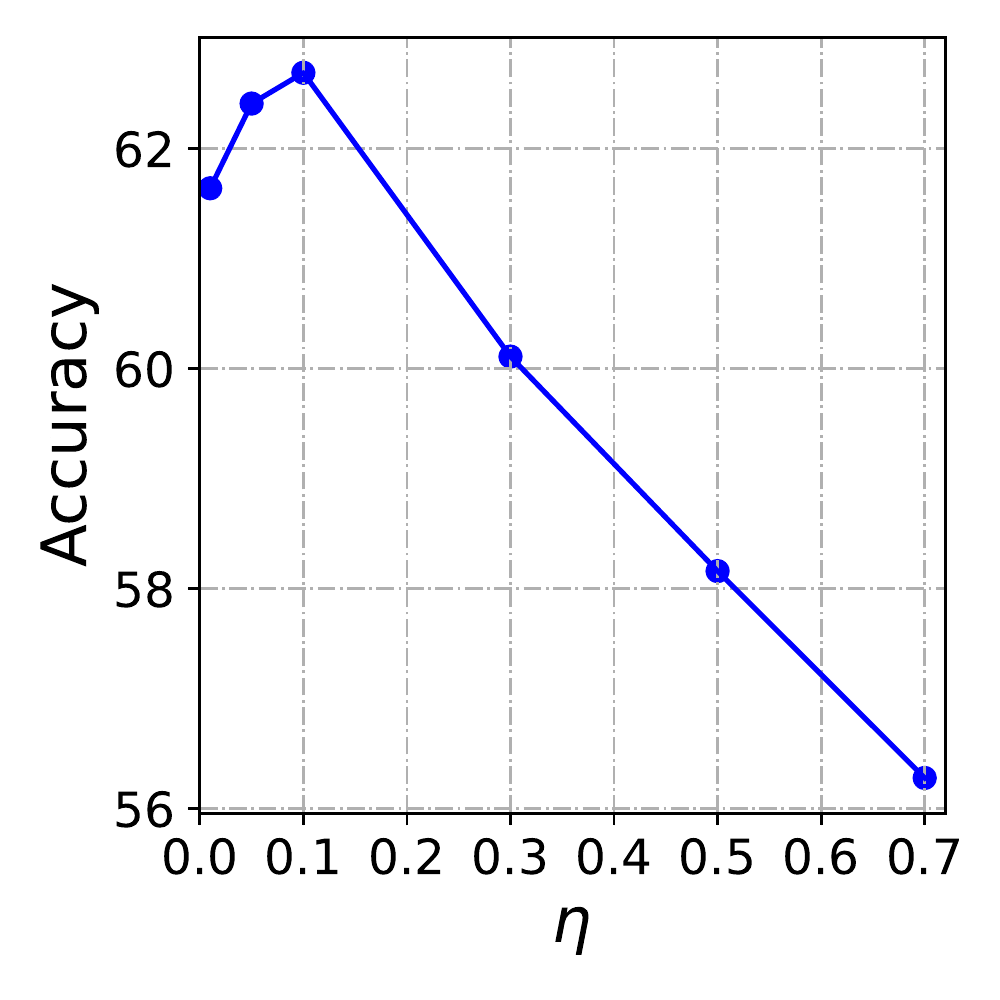}
                   \caption{$\eta$-accuracy}\label{f:etaacc}
        \end{subfigure}
        \begin{subfigure}[t]{0.32\linewidth}
            \centering
                   \footnotesize
                   \includegraphics[width=\textwidth]{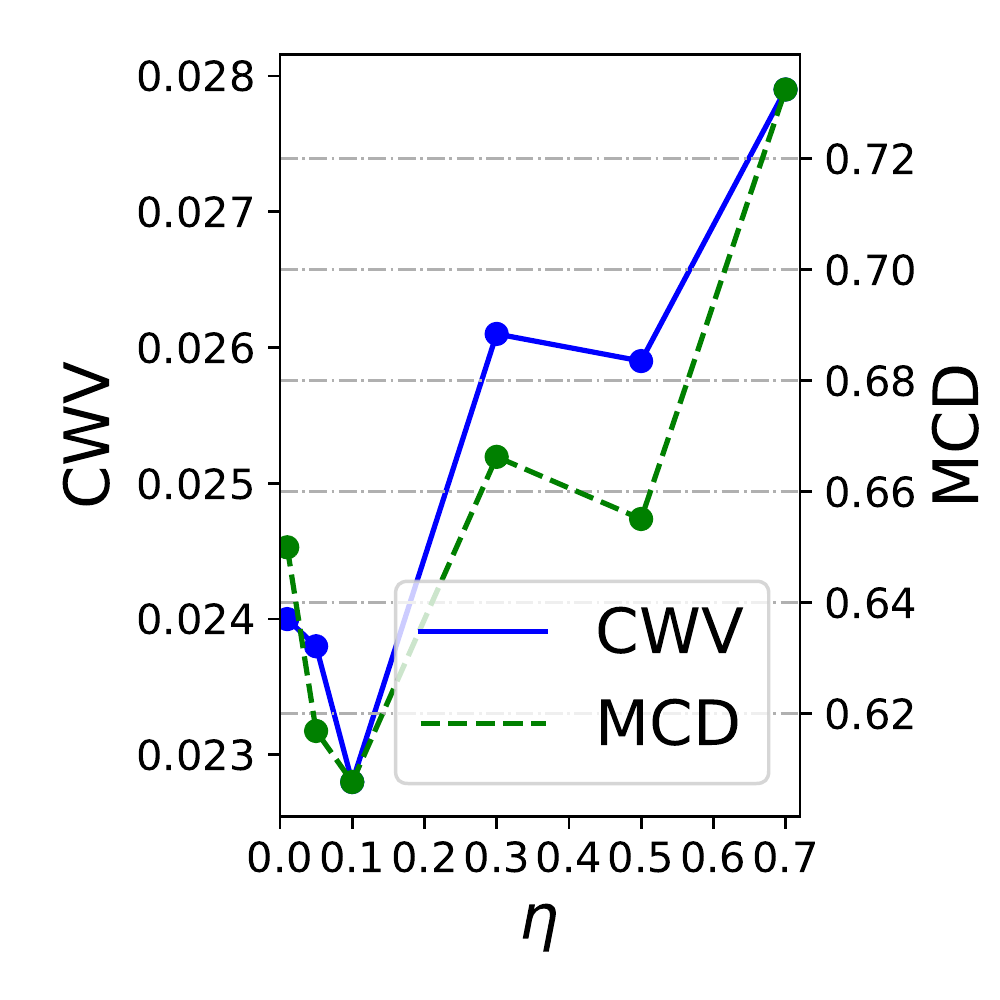}
                   \caption{$\eta$-CWV\&MCD}\label{f:etacwvmcd}
        \end{subfigure}
        \begin{subfigure}[t]{0.32\linewidth}
            \centering     
            \footnotesize
            \includegraphics[width=\textwidth]{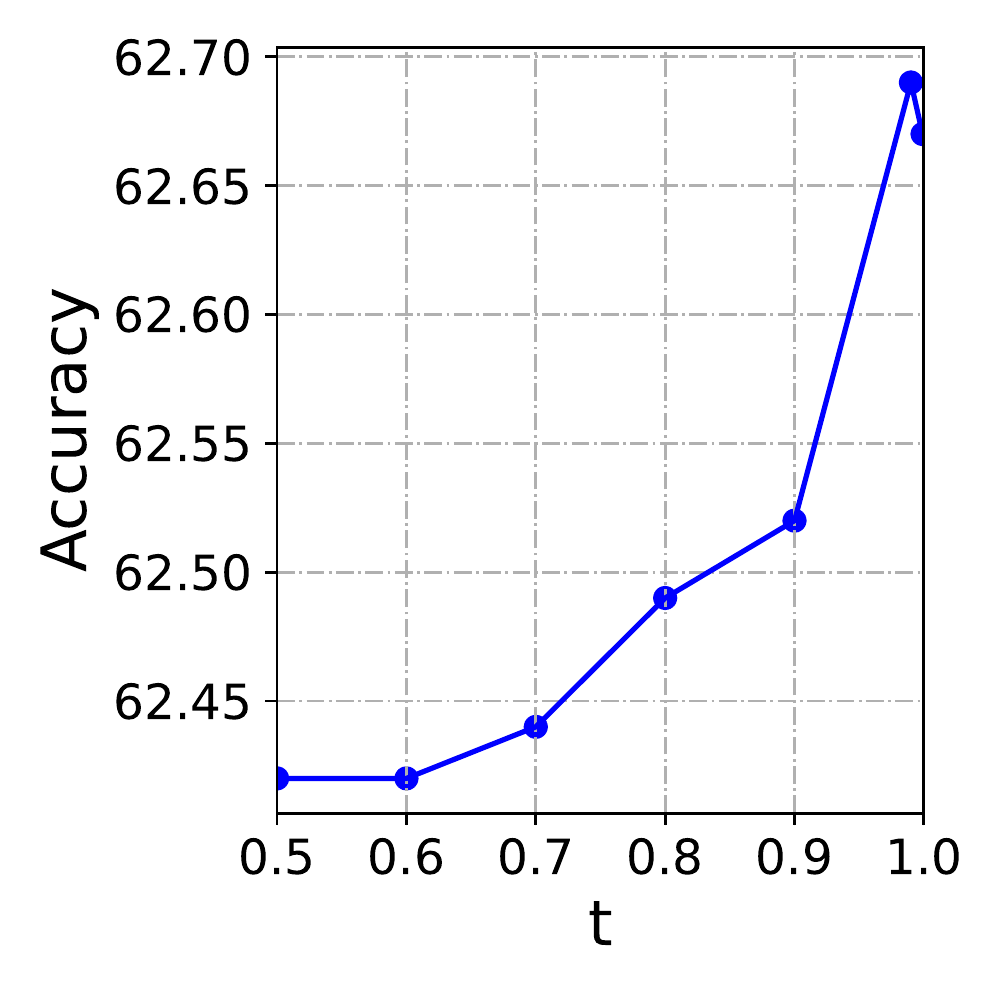}
                   \caption{$t$-accuracy}\label{f:tacc}
        \end{subfigure}
        
        \begin{subfigure}[t]{0.32\linewidth}
            \centering
                   \footnotesize
                   \includegraphics[width=\textwidth]{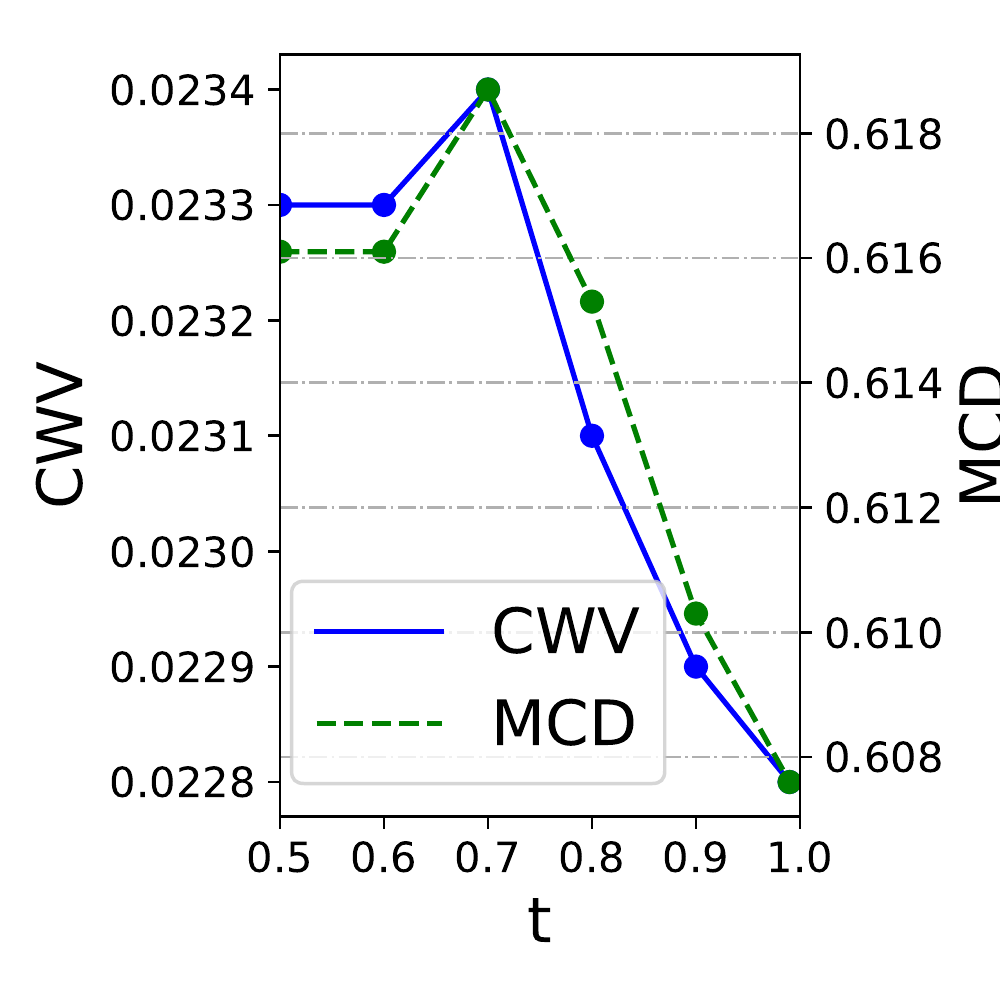}
                   \caption{$t$-CWV\&MCD}\label{f:tcwvmcd}
        \end{subfigure}
        \begin{subfigure}[t]{0.32\linewidth}
            \centering     
            \footnotesize
            \includegraphics[width=\textwidth]{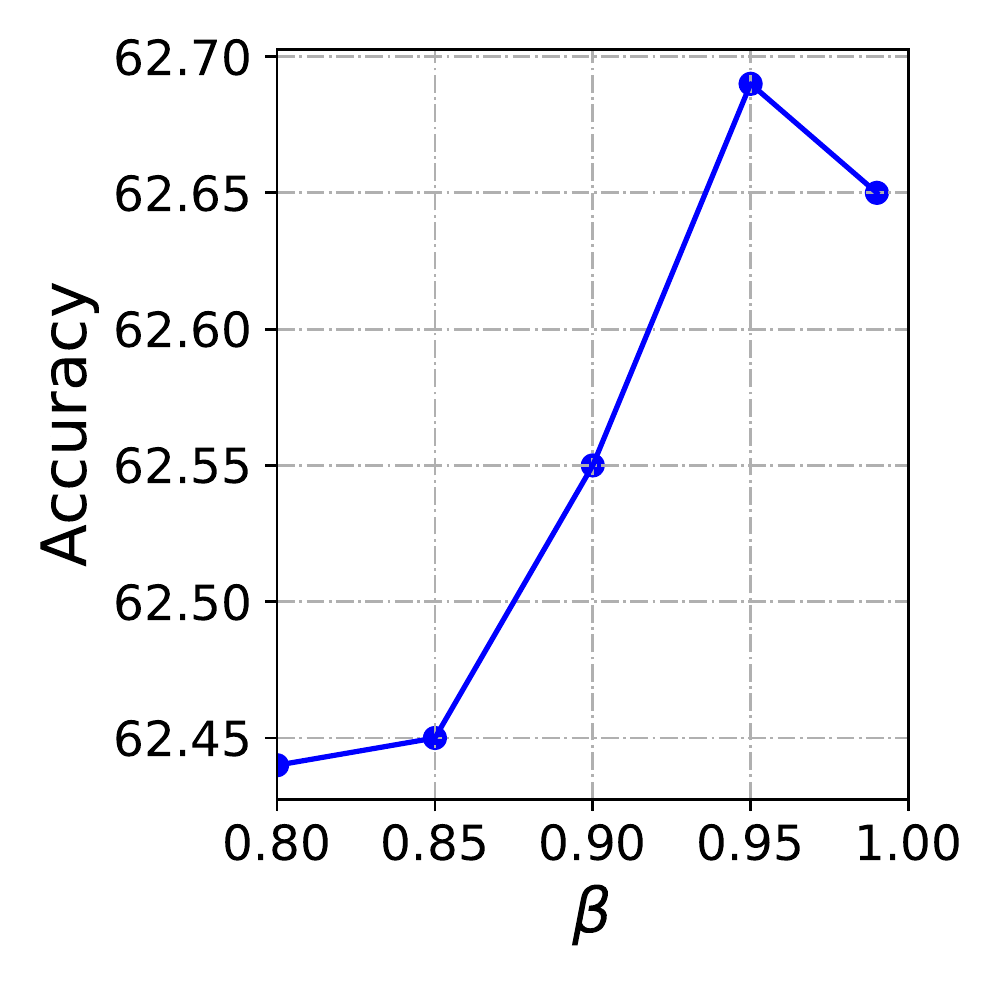}
                   \caption{$\beta$-accuracy}\label{f:betaacc}
        \end{subfigure}
        \begin{subfigure}[t]{0.32\linewidth}
            \centering
                   \footnotesize
                   \includegraphics[width=\textwidth]{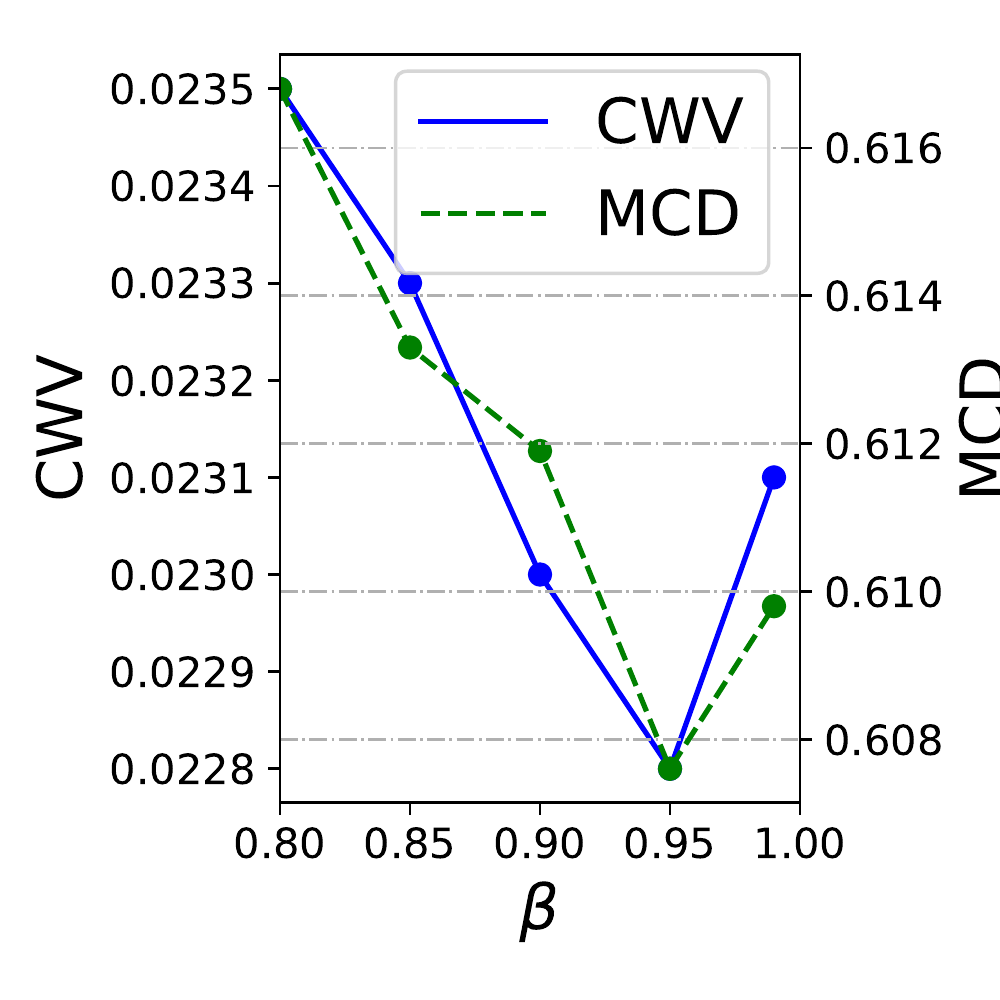}
                   \caption{$\beta$-CWV\&MCD}\label{f:betacwvmcd}
        \end{subfigure}

    \caption{Effect of $\eta$, $t$ and $\beta$ on model average performance.}\label{fig:hyperparameter}
    \vspace{-10pt}
\end{figure}

\begin{table}[]
    \centering
    \caption{Comparison among different divergence metrics used in dataset refinement.}
    \label{tab:divmetric}
    \scalebox{0.6}{
    \begin{tabular}{@{}lrrrrrr@{}}
    \toprule
    Metric & Avg Acc.  & Avg CWV & Avg MCD & Last Acc.  & Last CWV & Last MCD \\ 
    \midrule
    Jensen-Shannon  & 62.69    & 0.0228           & 0.6076     & 56.60    & 0.0256           & 0.6888       \\ 
    Kullback-Leibler  & 60.74  & 0.0239             & 0.6297   & 54.32    & 0.0262           & 0.7206          \\ 
    Hellinger  & 59.89  & 0.0252             & 0.6337          & 53.97    & 0.0282           & 0.7178  \\ 
    \bottomrule
    \end{tabular}
    }
    \vspace{-5pt}
\end{table}

\begin{figure}[h]
    \centering
    \scalebox{0.99}{
    \includegraphics[trim={0 70 0 70},clip,width=\linewidth]{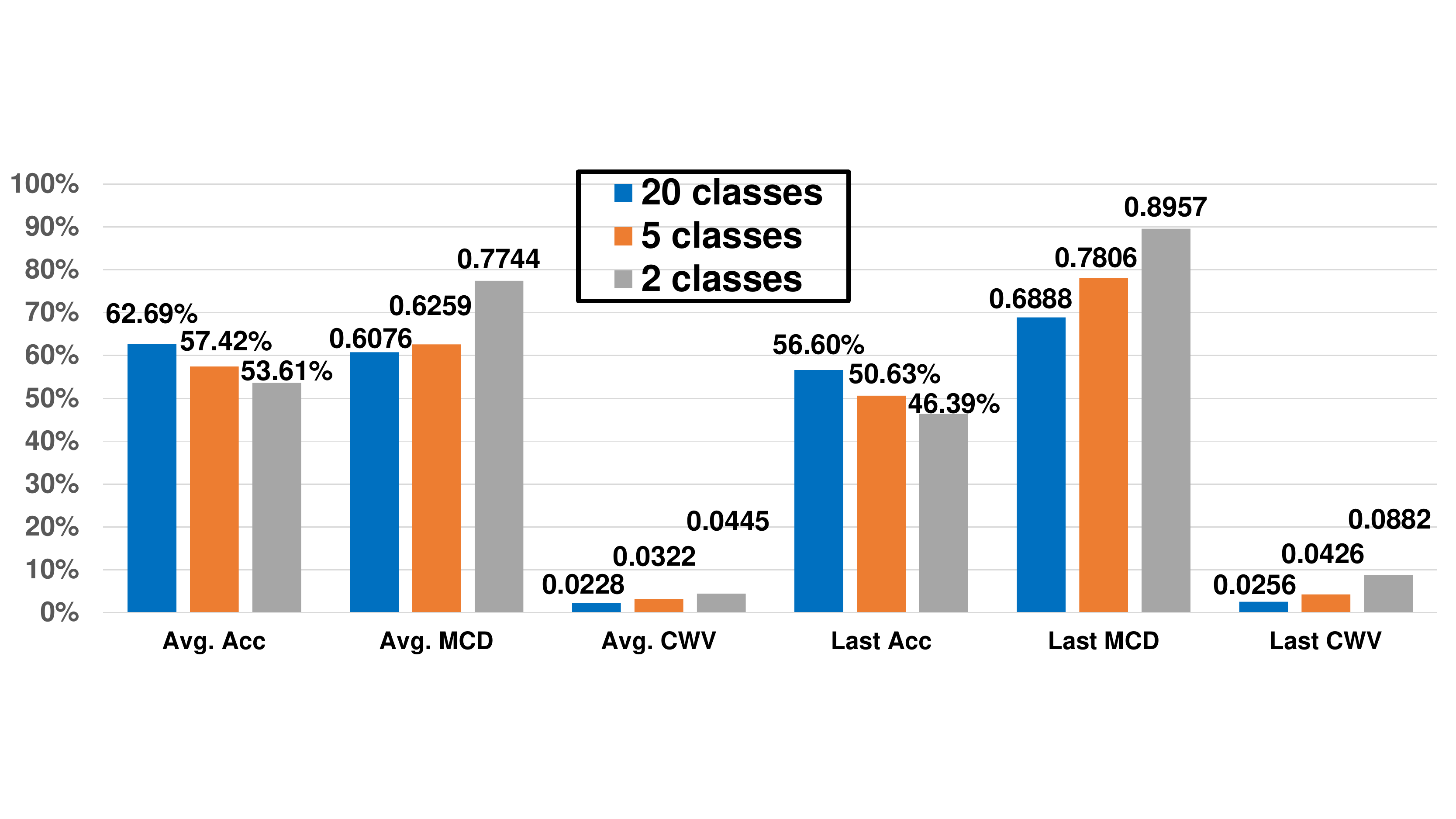}
    }
    \caption{Comparison among different split of class number.}\label{fig:classnum}
    \vspace{-10pt}
\end{figure}

\noindent \textbf{Experiment Design:}
The implementation of \sys requires predetermining the hyperparameters and the metric used for dataset refinement.
In this section, we investigate how these configurations affect the fixing performance.
To this end, 
1) we conduct a comparison experiment for verification hyperparameters \(t\) and \(\beta\) varying between the interval \([0.5,1.0]\) and \([0.9,1.0]\), respectively;
2) we conduct a comparison experiment for dataset refinement hyperparameter $\eta$ varying between the interval $[0.01,0.7]$;
3) besides the Jensen-Shannon divergence used by \sys to refine datasets, we also evaluate the performance using other metrics to measure the similarity of output, including Kullback-Leibler divergence and Hellinger distance.
4) we conduct a comparison experiment for different incremental class number~(2, 5, 20) on CIFAR-100;
We follow the CIL setting in \autoref{sec:rq1} and conduct experiments on the CIFAR-100 dataset.
We compare these methods in terms of the average performance~(except the base model) and model performance obtained in the last step.

\noindent \textbf{Results:}
We investigate the effect of hyperparameters $\eta$, \(t\) and \(\beta\), respectively.
\autoref{fig:hyperparameter} shows how hyperparameter assignment affects the performance of the model, which reports the average results.
Besides, we also study the effect of the choice of dataset refinement metric.
\autoref{tab:divmetric} reports the results with different dataset refinement metrics.

\textit{Effect of Hyperparameter \(\eta\):}
\sys leverages a configurable hyperparameter $\eta$ to control the degree of dataset refinement.
$\eta$ represents the threshold of dataset division.
As its value increases, more samples are classified as impurity samples, resulting in a smaller refined dataset.
As we can see from the~\autoref{fig:hyperparameter}, the model's performance first improves and then degrades as $\eta$ rises. 
\sys performs best when $\eta = 0.01$, so we choose this value as the default setting.

\textit{Effect of Hyperparameter \(t\) and \(\beta\):}
\sys leverages two configurable hyperparameters, \(t\) and \(\beta\), to adjust the strictness of the neuron coverage verification process.
\(t\) represents the activation threshold of a single neuron, and \(\beta\) represents the neuron coverage threshold of the model.
As we can see from the~\autoref{fig:hyperparameter}, as \(t\) and \(\beta\) increase, the model's performance also first improves and then degrades and performs best when $t = 0.99$ and $\beta = 0.95$.
So we choose these values as the default setting.

\textit{Effect of Dataset Refinement Metric:}
From~\autoref{tab:divmetric}, we can observe that the model using Jensen-Shannon divergence surpasses the Kullback-Leibler one by 2.75\%, 0.42\%, and 2.05\%, while surpassing the Hellinger one by 4.21\%, 5.56\%, and 2.67\% in accuracy, CWV, and MCD respectively.
It confirms that using the Jensen-Shannon divergence in \sys can achieve the best accuracy and fairness.

\textit{Effect of Incremental Class Number:}
To investigate the effect of class number, we split the dataset by different granularity. 
We split CIFAR-100 into batches of 20, 5, and 2 classes, corresponding to 5, 20, and 50 steps of incremental learning, respectively.
From~\autoref{fig:classnum}, we can observe that \sys with a split of 20 classes surpasses 5 classes one by 8.41\%, 41.23\%, and 7.46\%, while surpassing the 2 classes one by 14.48\%, 95.18\% and 27.45\% in accuracy, CWV, and MCD respectively.
It demonstrates that the more steps taken for incremental learning, the more serious degradation of the model performance.

\noindent \textbf{Analysis:}
Firstly, it can be seen from \autoref{fig:hyperparameter} that the performance of the model first increases and then drops as hyperparameters grow, which means that there are extreme points.
Since \(\eta\) is a hyperparameter that controls the size of the high-quality dataset, we can say that the performance of the high-quality dataset is not positively related to its size.
In our opinion, if the size of the high-quality dataset is too small, it will be difficult to including all important features; if it is too large, it will introduce more noise and disturbance of samples, which makes it difficult for model to extract the truly important features.
Similarly, if the values of t and b are too large, then the constraints of verification will be too strict, which results in insufficient data sampling; otherwise, it may lead to insufficient feature representation.

Last, \autoref{tab:divmetric} demonstrates that the performance of the model is directly impacted by the choice of the divergence metrics.
The Hellinger model performed the worst, which we believe is because it uses a lot of square and square root operations, which may lead to numerical issues, to calculate the Hellinger distance.
The Jensen-Shannon divergence outperforms the Kullback-Leibler model in terms of performance. 
We believe the reason is that the Kullback-Leibler divergence calculation is asymmetric, i.e., \(KL(P,Q)\) is not equal to \(KL(Q,P)\), which makes it difficult for the model to precisely measure the importance of samples on training.

\noindent \textbf{Accuracy-fairness trade-off:}
The model accuracy-fairness trade-off~\cite{kamiranDataPreprocessingTechniques2012} refers to the fact that improving the accuracy of a machine learning model on a given task may come at the cost of reduced fairness or equity in its predictions. 
In many cases, models trained to optimize accuracy learn to rely on discriminatory features in the data, such as race or gender, and make biased predictions. 
Conversely, attempts to ensure fairness or equity in a model's predictions may result in reduced accuracy, as the model may be forced to ignore or down-weight relevant features.
Notice that the trade-off may not exist in scenarios with no such conflicts~\cite{fishConfidenceBasedApproachBalancing2016,zafarFairnessDisparateTreatment2017,wickUnlockingFairnessTradeoff2019}.
This can occur when the relevant features for a task are equally distributed across different demographic groups, such as in some cases where the distribution of features is not biased toward any particular group.
In these cases, a model trained to optimize accuracy on a given task may also be fair since the most relevant features are not discriminatory.
We study if such a trade-off exists in \sys.
The value of \(\eta\), i.e., the degree of the dataset refinement, constraints the fairness requirement.
In extreme cases, without imposing fairness constraint~(\(\eta=0\)), \sys regards all samples as high-quality samples and vice versa.
Therefore, we study the accuracy-fairness trade-off by analzing the relationship between accuracy and \(\eta\).
Figure 11~(a)\&(b) show that the accuracy and fairness have the same trend (i.e., first increase and then drop as overfitting happens) as \(\eta\) grows, using CIFAR-100 dataset and ResNet-32 model.
Experiments on Flowers and Cars show similar results.
Notice that the trade-off effects depend on different settings (e.g., datasets, models), and results in other settings may be different.

\vspace{-10pt}
\subsection{RQ4: Insights for Incremental Learning}\label{sec:rq4}

Based on the study and discussion presented above, we can make some recommendations for enhancing performance of CIL models.

\textit{Increase the size of the sampled dataset.}
As we previously explained in \autoref{sec:biasdataset}, it is essential to keep more old samples since this can significantly lessen dataset bias.
Our observations indicate that the performance improvement of the model for the CIFAR-100 dataset is not readily apparent once the number of old samples is raised to 5000, which is 10\% of the total training sample size.
Specifically, the model utility differs from the best utility~(storing all old samples) by no more than 2\% when the number is 5000, while the performance from 500 to 5000 improves by nearly 10\% as a comparison.
This suggests that increasing the size of the sampled dataset can help improve model performance, but there is a marginal effect.

\textit{Maintain model class balance.}
A significant disparity in data amount across the model's classes has impact on its fairness performance~(as \autoref{fig:Imbalance} shown).
Therefore, it is necessary to preprocess the training dataset before training to keep the model class balance.

\textit{Force the model to learn important features.}
Features occupy the most important part of traditional machine learning fairness research, and for image classification systems, features still play an important role.
Throughout the training phase, the model must continually extract the data's features.
If the model misses certain crucial features during training, its effectiveness can be significantly diminished. 
It is possibly cased by the difficulties to extract certain features from the training dataset~(as \autoref{fig:mask} shown), or inadequate learning over these features~(as \autoref{fig:algbias} and \autoref{tab:ablation} shown).

Empirically, applying these strategies is beneficial to improve the fairness performance and utility of CIL models.
Developers, in our opinion, may obtain a better incremental learning pipeline by appropriately using the aforementioned training strategies.
\subsection{RQ5: Efficiency of \sys}\label{sec:rq5}

We measure the time of IL training by BiC, WA, iCaRL, and \sys on the same setting used in \autoref{sec:rq1} to evaluate the efficiency of different methods.
The results are presented in \autoref{tab:efficiency}.
The first column lists the three datasets and the remaining columns list the time costs of different training methods.
On average, \sys spends 2.2\% more time than BiC, and 6.7\% and 1.7\% less than WA and iCaRL.
Overall, the time spent is not significantly different from other methods, while \sys achieves better fixing performance.

\begin{table}[]
    \centering
    \caption{Time to train a model.}
    \label{tab:efficiency}
    \scalebox{0.75}{
    \begin{tabular}{@{}lrrrrrr@{}}
    \toprule
    Dataset & \sys  & BiC & WA & iCaRL  \\ 
    \midrule
    CIFAR-100  & 53784s    & 52386s   & 60377s    & 55347s   \\ 
    Flowers  & 12835s  & 12390s    & 13659s  & 13031s     \\ 
    Cars  & 8446s  & 8403s      & 8576s     & 8498s    \\ 
    ImageNet  & 215873s  & 214896s      & 220089s     & 217831s    \\ 
    \bottomrule
    \end{tabular}
    }
    \vspace{-15pt}
\end{table}

\section{Threat to Validity}\label{sec:threat}
\sys is currently evaluated on 3 datasets, which may be limited.
Similarly, there are configurable parameters used in \sys, and even though our experiments show that they are good enough to achieve high fixing results, this may not hold when the size of model is significantly larger or smaller.
To mitigate these threats, all the original and repaired training scripts, model architecture and training configuration details, implementation including dependencies, and evaluation data are available at \cite{AnonymizedRepositoryAnonymousa} for reproduction.

\vspace{-5pt}

\section{Related work \& Discussion}\label{sec:rw}

\noindent
{\bf Class Incremental Learning.}
Class incremental learning becomes one active topic recently.
Several works~\cite{ostapenkoLearningRememberSynaptic2019,yuSemanticDriftCompensation2020} attempt to train models without access to previously seen data, but the performance is not ideal.
Prevalent strategies, which can primarily be examined through representation learning and classifier learning, are based on the rehearsal method with limited data memory.

\textit{Representation Learning.}
The following three categories are generally used to classify contemporary works.
Regularization-based methods~\cite{kirkpatrickOvercomingCatastrophicForgetting2017,zenkeContinualLearningSynaptic2017,leeOvercomingCatastrophicForgetting2017,aljundiMemoryAwareSynapses2018,chaudhryRiemannianWalkIncremental2018} estimate changes of key parameters, and then update the posterior of model parameters sequentially. 
Their computation, however, frequently calls for approximations with a strong model premise.
Distillation-based methods~\cite{rebuffiIcarlIncrementalClassifier2017,zhaoMaintainingDiscriminationFairness2020,wuLargeScaleIncremental2019,castroEndtoendIncrementalLearning2018,douillardPodnetPooledOutputs2020,houLearningUnifiedClassifier2019} leverage knowledge distillation~\cite{hintonDistillingKnowledgeNeural2015a} to maintain the representation.
iCaRL~\cite{rebuffiIcarlIncrementalClassifier2017} and EE2L~\cite{castroEndtoendIncrementalLearning2018} compute the distillation loss on the network outputs.
Instead of using the network prediction, UCIR~\cite{houLearningUnifiedClassifier2019} applies the distillation loss using normalized feature vectors. 
PODNet~\cite{douillardPodnetPooledOutputs2020} limits the change of model by using a spatial-based distillation loss. 
Structure-based methods~\cite{hungCompactingPickingGrowing2019,yanDynamicallyExpandableRepresentation2021} keep the learned parameters related to previous classes fixed and allocate new parameters in various ways, such as unused parameters or additional networks to learn new knowledge.

\textit{Classifier Learning.}
Due to memory constraints, the class imbalance problem is the main challenge for classifier learning methods.
Some studies, such as LwF.MC~\cite{liLearningForgetting2018} and RWalk~\cite{chaudhryRiemannianWalkIncremental2018}, train the extractor and classifier together in a single training session.

\noindent
{\bf Fairness of ML.}
Fairness issues in ML have drawn a lot of attention as a result of the expanding usage of automated decision-making methods and systems, such as standardized testing in higher education~\cite{clearyTestBiasValidity1966}, employment~\cite{guionEmploymentTestsDiscriminatory1966, raghavanMitigatingBiasAlgorithmic2020, vandenbroekHiringAlgorithmsEthnography2019}, and re-offense judgement~\cite{oneilWeaponsMathDestruction2016, brennanEmergenceMachineLearning2013, berkFairnessCriminalJustice2021, berkAccuracyFairnessJuvenile2019}.
Besides, governments (e.g. the EU~\cite{voigtEuGeneralData2017} and the US~\cite{presidentBigDataSeizing2014, presidentBigDataReport2016}), organizations~\cite{markhamEthicalDecisionmakingInternet2012}, and the media have asked for more public responsibility and social understanding of ML.

To address the concern above, fairness testing for ML models becomes an important research direction, and its approaches are mostly based on generation techniques. 
THEMIS considers group fairness using causal analysis and uses random test generation to evaluate fairness~\cite{angellThemisAutomaticallyTesting2018}.
AEQUITAS focuses on the individual discriminatory instances generation~\cite{udeshiAutomatedDirectedFairness2018}. 
Later, ADF combines global search and local search to systematically search the input space with the guidance of gradient~\cite{zhangWhiteboxFairnessTesting2020}. 
Symbolic Generation~(SG) integrates symbolic execution and local model explanation techniques to craft individual discriminatory instances~\cite{agarwalAutomatedTestGeneration2018}.

The ML model needs to be repaired after the fairness problem is found.
To mitigate dataset bias, pre-processing approaches are proposed, including correcting labels~\cite{kamiranClassifyingDiscriminating2009, zhangAchievingNonDiscriminationData2017}, revising attributes~\cite{feldmanCertifyingRemovingDisparate2015, kamiranDataPreprocessingTechniques2012}, generating non-discrimination data~\cite{sattigeriFairnessGANGenerating2019, xuFairganFairnessawareGenerative2018}, and obtaining fair data representations~\cite{beutelDataDecisionsTheoretical2017}.
To alleviate algorithm bias, many in-processing and post-processing approaches are proposed.
More specifically, these approaches apply fairness ~\cite{zafarFairnessConstraintsMechanisms2017,dworkFairnessAwareness2012}, propose an objective function considering the fairness of prediction~\cite{zhangMitigatingUnwantedBiases2018}, design a new training frameworks~\cite{xuFairganFairnessawareGenerative2018,adelOneNetworkAdversarialFairness2019,gaoFairneuronImprovingDeep2022}, or directly change the predictive labels of bias models' output~\cite{hardtEqualityOpportunitySupervised2016, pleissFairnessCalibration}.
%
Zheng et al. proposed NeuronFair, a DNN fairness testing framework that identifies biased neurons, generates discriminatory samples as seeds guided by these neurons, and perturbs seeds to generate more instances~\cite{zhengNeuronFairInterpretableWhiteBox2021}.
Linear-regression-based Training Data Debugging~(LTDD) is another fairness testing tool.
It focuses on detecting which data features and which parts of them are biased~\cite{liTrainingDataDebugging2022}.
Peng et al. proposed xFAIR, a model-based fairness fixing method, which mitigates bias and explains the cause by leveraging correlations among data features~\cite{pengFairMaskBetterFairness2022,pengXFAIRBetterFairness2022}.

\noindent
{\bf Discussion.}
Our method, \sys, is based on knowledge distillation and shares similarities with BiC and WA. 
However, unlike these methods, \sys does not introduce additional layers or classifiers to the model, as this can negatively impact the model's generalizability. 
While BiC and WA aim to enhance the accuracy of IL, \sys focuses on improving fairness without significantly sacrificing accuracy. 
Currently, no other IL methods evaluate and constrain the fairness metrics of the model, making \sys the first to address fairness issues in the IL model. 
We aim to raise awareness of fairness issues in IL systems and mitigate potential negative societal impacts. 
Compared to BiC and WA, \sys achieves comparable accuracy performance while significantly improving fairness performance. 
\sys has a higher computational cost than BiC but lower than WA, and we plan to explore ways to improve its efficiency in future work. 
In summary, \sys is distinct from existing methods in that it does not add extra layers and systematically evaluates the fairness of IL models, making it a stronger performer with further potential for optimization in efficiency.
Vision transformer~(ViT) has recently been an emerging visual model.
Limited by hardware, we did not test our method on ViT, but we believe dataset refinement can still be applied since the approach does not depend on the model architectures.
It will be interesting future work to apply our method to ViTs.

\vspace{-5pt}
\section{Conclusion}\label{s:conclusion}

Inspired by software debugging, we propose and develop \sys, an automated class-based incremental learning model fairness debugging technique powered by dataset and training refinement.
It can identify important samples and train the model using the debiased training method on these samples.
Our evaluation results show that \sys construct high-quality datasets that effectively fix model fairness bugs in class-based incremental learning.

\bibliographystyle{ACM-Reference-Format}
\bibliography{CILIATE}

\end{document}